\documentclass[11pt]{article}
\usepackage{float}
\usepackage{graphicx}
\usepackage{booktabs}      
\usepackage{multirow}      
\usepackage{adjustbox}     
\usepackage{array}         
\usepackage{caption}       

\usepackage[most]{tcolorbox}
\usepackage{xcolor}
\usepackage{listings}
\usepackage{titlesec}

\titlespacing*{\section}{0pt}{10pt}{5pt}
\titlespacing*{\subsection}{0pt}{8pt}{4pt}
\titlespacing*{\subsubsection}{0pt}{6pt}{3pt}

\definecolor{PromptTitle}{RGB}{55,55,55}
\definecolor{PromptBack}{RGB}{247,247,247}
\definecolor{PromptFrame}{RGB}{85,85,85}

\lstdefinestyle{promptstyle}{
  basicstyle=\ttfamily\scriptsize\color{black},
  breaklines=true,
  breakatwhitespace=false,
  columns=fullflexible,
  keepspaces=true,
  showstringspaces=false,
  tabsize=2
}

\newtcblisting{promptbox}[1]{
  listing only,
  breakable,
  enhanced,
  sharp corners,
  boxrule=0.6pt,
  colback=PromptBack,
  colframe=PromptFrame,
  coltitle=white,
  colbacktitle=PromptTitle,
  fonttitle=\bfseries\small,
  title=#1,
  left=4pt,
  right=4pt,
  top=3pt,
  bottom=3pt,
  before skip=4pt,
  after skip=5pt,
  listing options={style=promptstyle}
}

\usepackage{acl}
\usepackage{adjustbox}

\usepackage{times}
\usepackage{latexsym}

\usepackage[T1]{fontenc}

\usepackage[utf8]{inputenc}

\usepackage{microtype}

\usepackage{inconsolata}

\usepackage{graphicx}

\usepackage{amsmath} 
\usepackage{amssymb} 
\usepackage[ruled,vlined]{algorithm2e}

\usepackage{tabularx}

\usepackage{makecell}

\usepackage[utf8]{inputenc}

\usepackage{textgreek}

\title{CWF: A Collaborative Writing Framework for Personalized and Reliable Popular Science Writing}

\author{
  \textbf{Ruibiao Fu}\textsuperscript{1,\textdagger},
  \textbf{Di Tang}\textsuperscript{1,\textdagger},
  \textbf{Yunlong Yang}\textsuperscript{1,\textdagger},
  \textbf{Ran Wang}\textsuperscript{1,2,*},
  \textbf{Sicheng Lu}\textsuperscript{1} \\
  \textbf{Peixuan Wu}\textsuperscript{1},
  \textbf{Xiaoyu Fan}\textsuperscript{1},
  \textbf{Jiacheng Ma}\textsuperscript{1},
  \textbf{Haozhe Luo}\textsuperscript{1},
  \textbf{Yang Xiao}\textsuperscript{3} \\[-1pt]
  \parbox{0.98\textwidth}{
    \centering
    \textsuperscript{1}\,School of Journalism and Information Communication,
    Huazhong University of Science and Technology, Wuhan 430074, China \\
    \textsuperscript{2}\,Philosophy and Social Sciences Laboratory of Big Data and National Communication Strategy,
    Ministry of Education, Wuhan 430074, China \\
    \textsuperscript{3}\,The National Key Laboratory of Multispectral Information Intelligent Processing Technology,\\
    School of Artificial Intelligence and Automation,
    Huazhong University of Science and Technology, Wuhan 430074, China \\[3pt]
    \texttt{\{furuibiao3,tangdi030223,yunlongyang64,xiaoyuf30\}@gmail.com} \\
    \texttt{\{rex\_wang,loo\_seychan,w\_px,jiacheng\_ma,luo\_haozhe,Yang\_Xiao\}@hust.edu.cn} \\
  }
}

\begin{document}
\setlength{\titlebox}{18\baselineskip}
\maketitle
\begingroup
\renewcommand{\thefootnote}{\fnsymbol{footnote}}
\footnotetext[2]{The three authors contribute equally to this work.}
\footnotetext[1]{Corresponding author.}
\endgroup
\begin{abstract}

We introduce \textbf{Personalized and Reliable Popular Science Writing}, a novel task that requires adapting scientific explanations to audiences with different cognitive levels while preserving factual accuracy. However, improving personalization often introduces simplifications that increase the risk of hallucination and factual distortion.
To address these challenges, we first construct a dataset of 39,134 entries and a reader-centric \textbf{Personalized Science Communication Benchmark (PSCB)} that jointly evaluates audience adaptation and factual accuracy.
To reduce data and computational requirements while improving generalization across domains and audiences, we introduce DA-MoE, which explicitly decouples audience adaptation from domain knowledge through separate modeling.
To enable robust verification and revision in evidence-scarce scenarios, a multi-agent fact-checking mechanism that augments limited evidence with role-specific agent debate and propagates confidence over a graph is proposed.
Experiments on PSCB show that our approach achieves state-of-the-art performance. Our code is open-sourced at https://github.com/DPInnovationWorks/CWF.
\end{abstract}

\section{Introduction}

\begin{figure}
    \centering
    \includegraphics[width=1\linewidth]{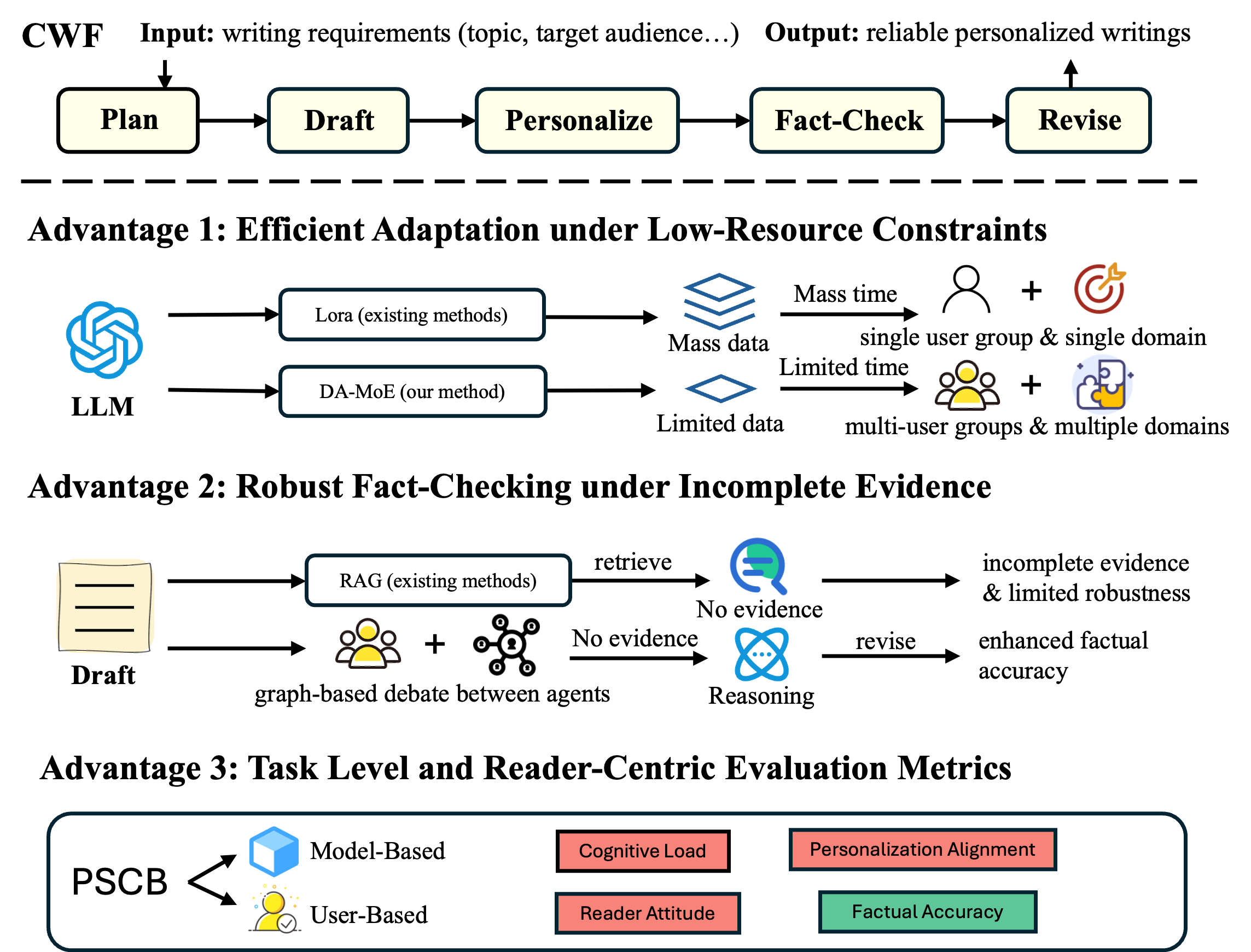}
    \caption{Comparison between Specialized Approaches (Fine-tuning and RAG-based methods) and Collaborative Writing Framework (CWF).}
    \label{fig:overall_contrast}
\end{figure}

Popular science writing aims to communicate scientific knowledge to audiences in an accurate and accessible way. Large language models (LLMs) have introduced new possibilities for this task by enabling flexible text generation across domains and writing styles. However, effective popular science writing requires more than fluent generation. The generated article should be adapted to the target audience in both writing style and cognitive adaptation, while remaining factual accuracy. Existing methods mainly rely on prompt engineering \cite{kim2024steering, tang2024step}, retrieval-based verification \cite{min-etal-2023-factscore}, or supervised fine-tuning (SFT) \cite{goldsack-etal-2022-making}. Despite their effectiveness, these approaches usually treat writing style, audience adaptation, and factual accuracy as separate objectives, making it difficult to generate articles that are both personalized and reliable.

This limitation becomes more pronounced in SFT. High-quality popular science writing often requires jointly aligned data that couples domain knowledge, audience adaptation, and writing style \cite{guo2022automatedlaylanguagesummarization}. Such data is costly to collect because it must contain both accurate knowledge and audience-adaptive expressions. Moreover, as a generative task, popular science writing may produce factual omissions and hallucinated explanations. Therefore, the key challenge is not only to reduce the dependence on large-scale paired data for domain and audience adaptation, but also to ensure factual accuracy throughout the generation process.

Existing benchmarks do not adequately address this challenge, they typically emphasize stylistic preference \citep{august2024know, tang2024step} or factual accuracy in popular science writing \citep{min2023factscore, damm2024wispermed, plainqafact2025}, but rarely consider both of them within a unified benchmark. Audience adaptation often changes explanation granularity, analogy use, implicit reasoning, and conceptual simplification, which may increase the risk of unsupported information or factual distortion \citep{devaraj-etal-2022-evaluating, taylor2018problem}. Accordingly, PSCB benchmark  is introduced for personalized and reliable popular science writing, as shown in Figure~\ref{fig:overall_contrast}. PSCB jointly evaluates cognitive adaptation and factual accuracy. Beyond reader attitude and personalization alignment, PSCB first introduces Cognitive Load \citep{sweller1988cognitive, salemi2023lamp, ouwehand2021measuring, zu2021subjective} as a metric and decomposes metrics into fine-grained dimensions for assessing cognitive adaptation beyond surface style. 

To address the challenges of costly jointly aligned data and the difficulty of balancing personalization with factual accuracy, we design the Collaborative Writing Framework (CWF). We first introduce a decoupled MoE architecture for popular science writing. This framework relaxes the requirement for paired data that contains the target audience and scientific knowledge. This design allows us to construct a dataset of 39,134 entries that does not rely on joint alignment between domains and audiences, enabling training across different scientific domains and reader groups. However, simple MoE faces difficulty when transferring to new domains, since generating articles in different domains may require retraining the whole model. We propose DA-MoE, whose key idea is to assign adaptation tasks to individual pluggable experts and to separate audience-specific adaptation from domain-specific knowledge, so that cognitive adaptation and knowledge can be learned from different sources. This allows to adapt to new domains with only lightweight training.

Popular science writing requires more than cognitive adaptation. Even well-personalized generations may contain omissions, unsupported claims, or hallucination, making factual verification essential. CWF therefore incorporates a fact-checking module to preserve factual accuracy during generation. However, RAG-based methods struggle when direct evidence is sparse or conflicting because they rely primarily on retrieved evidence rather than deriving additional support by reasoning over indirect viewpoints or evidences \citep{asai2023self}. Moreover, most post-hoc fact-checking methods rely on direct model judgments over completed outputs, not only producing unquantifiable results but also cannot guide the writing process \citep{jiang-etal-2023-active, min-etal-2023-factscore, NEURIPS2023_91f18a12, wei2024long}. Furtherly, we propose an verification mechanism that combines multi-agent debate with graph-based inference. Specifically, multi-agent debate simulates an interactive review process, and helps validate claims by introducing complementary perspectives and broadening retrieval angles when direct evidence is incomplete. Intervention applies verification during the generation process rather than only after generation, allowing multi-agent factual feedback to guide the writing process itself. Finally, graph-based inference connects claims, retrieved evidence, and agent reasoning, enabling additional evidence to be inferred from indirect viewpoint while making factual judgments traceable rather than black-box decisions.

In summary, \textbf{the main contributions} of this work are as follows:
\begin{itemize}
    \item Personalized and reliable popular science writing task is proposed for the first time. We further introduce PSCB, a human-aligned benchmark that jointly evaluates audience adaptation and factual accuracy.
    \item We propose a DA-MoE framework for popular science writing that decouples domain knowledge from audience adaptation, reduces reliance on paired training data, and enables low-cost generalization to new domains.
    \item We propose a multi-agent fact-checking mechanism with graph-based reasoning. It simulates the debate process to support multi-perspective verification, intervention during generation, and quantifiable confidence scores for robust claim validation.
\end{itemize}

\section{Related Works}

\textbf{Popular Science Writing.}
Recent work in computational linguistics studied how to transform scientific content into language that general audiences can understand. Goldsack et al. \cite{goldsack-etal-2022-making} and Cheng et al. \cite{cheng2025vtechagp} formulate this problem as lay summarization or academic-to-public paraphrasing by constructing paired datasets and benchmarks, including PLOS, eLife, and VTechAGP. Fang et al. \cite{fang2024understanding} and Jiang et al. \cite{jiang2025jre} further study reliability in popular science generation by analyzing factual errors in LLM-generated biomedical summaries or using multi-agent interaction to improve science journalism generation. However, existing studies usually address cognitive adaptation and factual accuracy separately. They do not provide a unified framework that jointly considers personalization for different readers and factual accuracy in popular science writing. To bridge this gap, we formulate personalized and reliable popular science writing and introduce PSCB to jointly evaluate reader adaptation and factual accuracy.

\textbf{Personalized and Controllable Writing.} Personalized generation primarily utilizes Few-Shot Prompting or SFT \citep{houlsby2019parameter, brown2020language, li-liang-2021-prefix}. However, prompt-based methods often struggle with long-form text, while fine-tuning imposes strict data requirements to model target styles effectively. This limitation is particularly pronounced when combining heterogeneous styles, which requires constructing costly datasets that jointly capture multiple stylistic patterns \citep{ju2024beyond}. To reduce this dependence, we propose DA-MoE, which decouples audience adaptation from domain knowledge and enables flexible composition across reader groups and scientific domains.

\textbf{Fact-checking in LLMs.} Traditionally, fact-checking has heavily depended on human evaluators, a process that is not only time-consuming and expensive but also prone to subjective biases \citep{honovich2022true, devaraj-etal-2022-evaluating, min-etal-2023-factscore}. Subsequently, approaches combining LLMs with RAG emerged to integrate external databases \citep{asai2023self}. However, they often struggle when direct evidence is sparse or conflicting, due to a lack of deep reasoning capabilities. Recent methods have proposed decomposing complex claims into atomic units \citep{min-etal-2023-factscore, wei2024long} or simulating multi-agent debates \citep{sun2025towards, MA2026130103}. By supplementing the original claims with additional context, these approaches enable judgments from multiple perspectives. However, they remain post-hoc and static, unable to instruct in the generation process proactively. Also, while simulated multi-agent debates introduce interactive reasoning, they rely on model judgments without providing rigorous metrics. To address these issues, we propose a verification framework. By combining multi-agent debate with graph, our method not only synthesizes diverse perspectives to handle incomplete evidence but also provides quantifiable confidence scores to individual reasoning nodes, ensuring a transparent verification process.

\textbf{Evaluation Benchmark.} Existing benchmarks either emphasize readability and audience accessibility in science communication \citep{august2024know, tang2024step} or focus on factual accuracy and factuality evaluation in generated texts \citep{min2023factscore, min-etal-2023-factscore, damm2024wispermed, plainqafact2025, wei2024long}. However, these two lines of work are usually studied separately, making them insufficient for evaluating popular science generation across different audiences. Our Personalized Science Communication Benchmark (PSCB) evaluates whether a generated science article is both reliable and adapted to its intended readers. In real-world science communication, an effective article should not only convey scientifically grounded knowledge, but also adjust its conceptual difficulty, explanation strategy, and communicative style to different reader groups. PSCB therefore contains two complementary evaluation branches: Factual Reliability and Audience Adaptation.

\begin{figure*}[t]
  \includegraphics[width=1.0\linewidth]{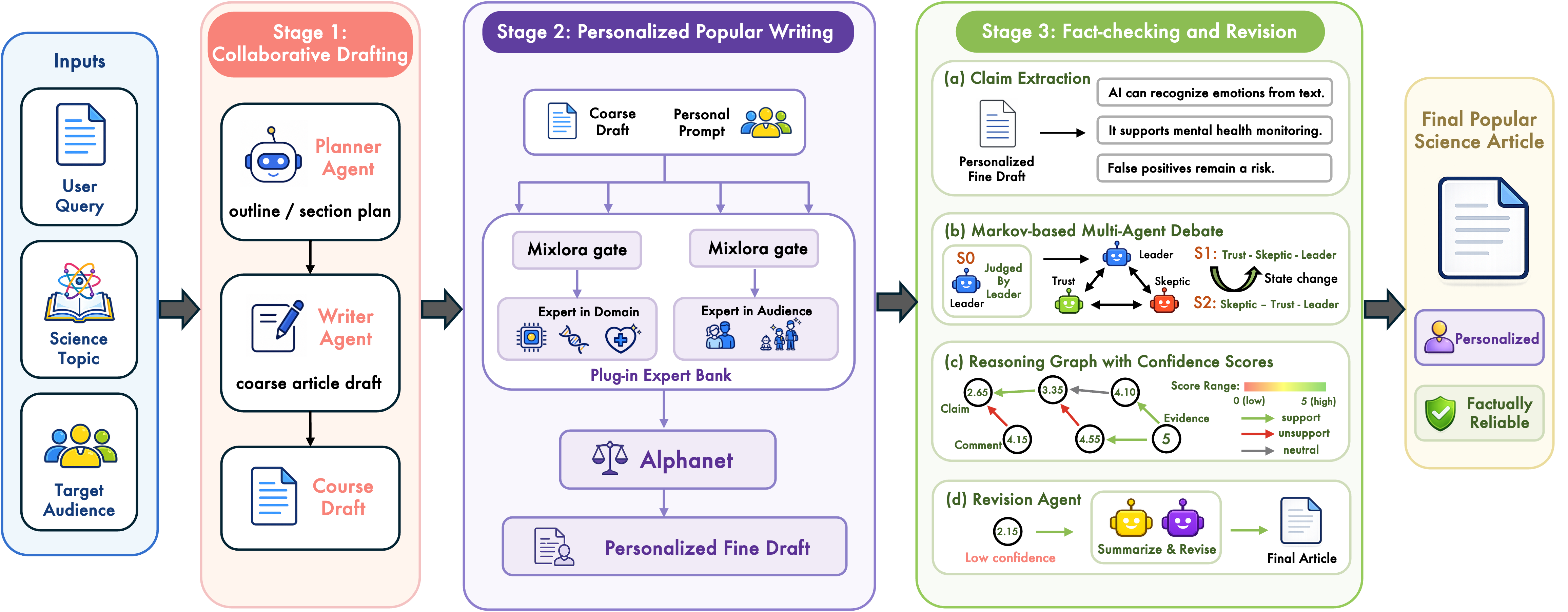}
  \caption {Overview of CWF, which integrates collaborative writing, personalized style modeling, and fact-checking with multi-agent discussion to enhance the quality and relevance of AI-generated science writings.}
  \label{fig:framework_overview}
\end{figure*}

\section{Collaborative Writing Framework}
As illustrated in Figure~\ref{fig:framework_overview}, the proposed CWF aims to generate coherent science writing that remain both personalized and factually reliable. CWF organizes the writing into three coordinated stages—drafting, personalization, and verification—to address the core challenge of balancing personalization and factual accuracy.

\subsection{Collaborative Writing}
Existing LLM-based popular science writing methods suffer from structural inconsistency, weak coherence, and limited control over narrative organization~\citep{shao-etal-2024-assisting,jiang-etal-2024-unknown}. To address this, we adopt a collaborative generation framework in which one agent plans the article structure and section summaries, while another expands them into fluent passages with examples and transitions. This process forms the initial draft and improves narrative coherence.

\subsection{DA-MoE}

To address the difficulty of data collection in SFT, we build a modular MoE-based generation framework on top of the plug-and-play MixLoRA architecture \cite{li2024mixlora}. The framework consists of two specialized MoE networks constructed from multiple Qwen2.5-3B backbone models: an audience-adaptation MoE for learning audience-specific writing styles, and a knowledge-domain MoE for learning topic-specific knowledge.

Formally, given an input \(x\), we construct two MoE modules: an audience-adaptation MoE and a knowledge-domain MoE. Let \(r \in \{s,k\}\) denote the expert type, where \(s\) represents audience adaptation and \(k\) represents knowledge domain. The corresponding expert sets are defined as:
\begin{equation}
\mathcal{E}^{r}=\{E^{r}_{1},E^{r}_{2},\ldots,E^{r}_{T_r}\},
\end{equation}
where \(T_s=M\) and \(T_k=N\). Each expert is initialized from Qwen2.5-3B. Audience experts are fine-tuned on data associated with specific audience adaptation, while knowledge experts are fine-tuned on data from specific scientific domains.

For each MoE module, a lightweight MixLoRA gating network assigns expert weights based on the input. The routing distribution is computed as:
\begin{equation}
g^{r}(x)=\mathrm{softmax}(\mathrm{MLP}_{r}(x)),
\end{equation}
where \(\mathrm{MLP}_{r}\) denotes the gating network for expert type \(r\). The output of each MoE module is then:
\begin{equation}
h^{r}(x)=\sum_{i=1}^{T_r} g^{r}_{i}(x) E^{r}_{i}(x),
\end{equation}
where \(g^{r}_{i}(x)\) is the routing weight assigned to the \(i\)-th expert in \(\mathcal{E}^{r}\).

The training process has three stages. First, each Qwen2.5-3B expert is independently fine-tuned on its assigned style or knowledge data, so that the style experts specialize in audience adaptation and the knowledge experts specialize in domain-specific scientific content. Second, after the expert parameters are fixed, we train the MixLoRA gating networks using a small amount of supervised data, allowing the model to learn how to route each input to appropriate style and knowledge experts. Third, we introduce an AlphaNet module to dynamically balance the contributions of the two MoE networks during generation.Specifically, AlphaNet is also parameterized as an MLP and directly predicts an input-dependent coefficient:
\begin{equation}
\alpha=\mathrm{MLP}_{alphanet}(x).
\end{equation}
The final hidden representation is then obtained by contrasting the style-oriented and knowledge-oriented MoE outputs:
\begin{equation}
h(x)=\alpha*h^{s}(x)+(1-\alpha)*h^{k}(x).
\end{equation}

To examine whether AlphaNet learns different balances for different target audiences, we analyze the distribution of $\alpha$ across reader groups.

\begin{table}[t]
\centering
\small
\resizebox{\columnwidth}{!}{%
\begin{tabular}{lccccc}
\toprule
\textbf{Reader Group} & \textbf{Mean} & \textbf{Std.} & \textbf{Median} & \textbf{Min} & \textbf{Max} \\
\midrule
Children  & 0.686 & 0.101 & 0.697 & 0.383 & 0.846 \\
Teenagers & 0.262 & 0.080 & 0.256 & 0.095 & 0.480 \\
Adults    & 0.183 & 0.059 & 0.176 & 0.079 & 0.351 \\
\bottomrule
\end{tabular}%
}
\caption{Distribution of $\alpha$ across different target reader groups ($n=100$ per group).}
\label{tab:alpha_distribution}
\end{table}

As shown in Table~\ref{tab:alpha_distribution}, the mean $\alpha$ decreases from children (0.686) to teenagers (0.262) and adults (0.183). Since a larger $\alpha$ indicates greater reliance on the style network, this result suggests that the model places more emphasis on style and readability for younger readers, while relying more on knowledge content for adult readers. This pattern is also consistent with previous studies showing that narrative and style-oriented presentation can improve comprehension and engagement for nonexpert audiences \cite{dahlstrom2014using,downs2014prescriptive}.

\begin{table}[t]
\centering
\footnotesize
\setlength{\tabcolsep}{4pt} 
\renewcommand{\arraystretch}{1.2} 
\begin{tabular}{l 
                >{\centering\arraybackslash}m{1cm} 
                >{\centering\arraybackslash}m{1cm} 
                >{\centering\arraybackslash}m{1.6cm} 
                >{\centering\arraybackslash}m{1.7cm}}
\toprule
\textbf{Method} & \textbf{Training Time} & \textbf{Data Amount} & \textbf{Data Type} & \textbf{Average PSCB score} \\
\midrule
LoRA & 2.1 hrs & 5000 & Dual-style & 2.985 \\
MoE & 8.3 hrs & 20000 & Single-style & 3.059 \\
DA-MoE & 22 min & 1000 & Single-style & 3.451 \\
\bottomrule
\end{tabular}
\caption{
Comparison of training time, data requirements, data type, and average PSCB score in different methods. 
}
\label{tab:training_time_data_half}
\end{table}

Table~\ref{tab:training_time_data_half} shows that our method achieves the highest average PSCB score while requiring substantially less training time and fewer training examples. This efficiency mainly comes from the modular design of our framework. When extending the model to a new domain, we do not need to collect large-scale dual-style data or retrain the full model. Instead, only a small amount of single-style data is required to train the gating network, while the existing style and knowledge experts can be reused.

\subsection{Fact-checking and Revision}

Our framework performs fact-checking through a multi-agent collaboration that simulates interactive discussions around scientific claims. Rather than relying on a single model to assess factuality, we decompose the verification process into a sequence of role-specific steps.

\textbf{Claim Extraction.}
Given a draft article, the system parses each paragraph into a set of atomic claims, ensuring that each statement can be independently verified. For each decomposed claim, we estimate its coverage to the original paragraph and the sums of all coverage scores are 1, which are used as weights in confidence aggregation.

\textbf{Markov Chain--Based Debate.}
We instantiate three complementary roles: Leader acts as an expert who gives the scientific judgment; Trust agent acts as a reader-oriented supporter who searches for reasonable evidence and complementary explanation; Skeptic agent acts as a critics who challenges unsupported claims and weak evidence. Science communication is not merely a one-way transmission of expert knowledge, but a mediated process involving expert judgment, audience-oriented interpretation, and critical gatekeeping \citep{wynne_1992, 2003Science, article00001}. Accordingly, the Leader, Trust, and Skeptic agents respectively operationalize these three functions for factual verification, with detailed role designs provided in Appendix~\ref{sec:appendix-prompt-templates}.

Given an atomic claim, the process starts from $S_0$, where the Leader provides an initial judgment. The subsequent debate order is controlled by the factuality signal $f(L_t)$ parsed from the latest Leader output, where $f(L_t)=1$ for supportive stances $(1,0.2)$ and $f(L_t)=0$ for neutral or opposing stances $(0,-0.2,-1)$:
\begin{equation}
\label{eq:markov-transition}
S_{t+1} =
\begin{cases}
S_2, & f(L_t)=1,\\
S_1, & f(L_t)=0.
\end{cases}
\end{equation}
When Leader initially supports a claim, Skeptic is invoked first to prevent premature acceptance; when Leader is uncertain or negative, Trust agent is invoked first to explore missing evidence or alternative explanatory paths. After each round, Leader updates the judgment based on both agents' arguments, and the transition rule is applied again.

We set the maximum number of debate rounds to $R_{\max}=3$, following the ablation results (provided in Appendix~\ref{sec:maximun_debate_rounds}) showing that three rounds provide the most stable confidence estimates. The debate stops early if all agents reach the same stance after one complete debate round; otherwise, it continues until $R_{\max}$ is reached.

\textbf{Confidence Calculation.}
The graph assigns every claim, agent statement, and evidence a confidence score on $[0,5]$. Evidence nodes are fixed at confidence $5$, while claim and agent-statement nodes start from a neutral prior of $2.5$. Each agent statement carries a stance value in $\{-1,-0.2,0,0.2,1\}$, indicating whether it strongly or weakly supports, opposes, or remains neutral toward its parent node. Confidence is propagated bottom-up through the graph instead of being decided by vote. For each statement, the system aggregates confidence by combining evidence and child statements, then maps the signed result back to the $[0,5]$. The final claim confidence is computed from the statements and direct evidence.

\textbf{Consensus Summarization and Writing Revision.}
The system produces a judgment based on the final reasoning graph and the calculated confidence scores. Claims with low confidence or detected inconsistencies will be revised by a summarizer agent who aggregates the discussion outcomes. These revised claims are then merged to form the verified article, ensuring that inaccurate content is revised prior to publication.

\section{PSCB Benchmark}
\label{sec:pscb}

\textbf{Dataset.}
PSCB dataset is constructed with 39,134 entries for training across scientific domains and reader groups. We collect popular science writing from science communication sources, including science-oriented WeChat public accounts and science books, covering three domains: AI, Biology, and Medicine. To model audience diversity, we further organize data for target reader groups: child, teen, and adult (Appendix~\ref{sec:appendix_participants}). 

\textbf{Metrics.}
As shown in Table~\ref{tab:evaluation_dimensions} and Appendix~\ref{sec:metrics_and_dimensions}, we score each article using three personalization metrics in 0--5 scale: Cognitive Load (CL), Personalization Alignment (PA), and Reader Attitude (RA). CL measures whether the explanation matches the reader's cognitive capacity \citep{sweller1988cognitive, salemi2023lamp, ouwehand2021measuring, zu2021subjective}; PA measures whether the content selection, knowledge level, linguistic style, and contextual framing are tailored to the target audience \citep{kreuter2003tailored, flek-2020-returning, moorjani-etal-2022-audience}; RA measures reader engagement, credibility, and continuance intention. \citep{obrien2008engagement, appelman2016message, bhattacherjee2001understanding}. By jointly modeling these three metrics, PSCB better evaluates both cognitive adaptation and domain knowledge.

\begin{table}[t]
\centering
\small
\setlength{\tabcolsep}{3pt}
\renewcommand{\arraystretch}{1.12}
\begin{adjustbox}{max width=\columnwidth}
\begin{tabular}{p{2.1cm}p{4.9cm}}
\toprule
\textbf{Metric} & \textbf{Dimensions} \\
\midrule
Cognitive Load 
&
\begin{tabular}[t]{@{}l@{}}
\textbf{INTR}: Intrinsic Fit \\
\textbf{EXTR}: Extraneous Burden Control \\
\textbf{GERM}: Germane Support
\end{tabular}
\\
\midrule
Personalization Alignment
&
\begin{tabular}[t]{@{}l@{}}
\textbf{CONT}: Content Relevance \\
\textbf{KNOW}: Knowledge-Level Fit \\
\textbf{STYLE}: Style Consistency \\
\textbf{CONTX}: Contextual Resonance
\end{tabular}
\\
\midrule
Reader Attitude
&
\begin{tabular}[t]{@{}l@{}}
\textbf{ENG}: Engagement Appeal \\
\textbf{TRU}: Trust and Credibility \\
\textbf{CONTI}: Continuance Intention
\end{tabular}
\\
\bottomrule
\end{tabular}
\end{adjustbox}
\caption{Personalization metrics and dimensions in PSCB. Each dimension is scored on a 0--5 scale by both LLM judges and human annotators.}
\label{tab:evaluation_dimensions}
\end{table}

\subsection{LLM-as-a-Judge in Personalization}

PSCB evaluates cognitive adaptation and domain knowledge through a unified protocol that combines dynamically weighted LLM-as-a-judge scoring, human questionnaires, and consistency analysis between the two evaluation sources.

For each generated article $y_i$, science domains or writing prompt $t_i$, and target reader persona $p_i$, PSCB evaluates three personalization-oriented metrics:
$\mathcal{M}=\{\mathrm{CL}, \mathrm{PA}, \mathrm{RA}\}$, as shown in Table~\ref{tab:evaluation_dimensions}.

\textbf{Dynamic Dimension Weighting.}
Before scoring the article, a meta-evaluator LLM analyzes only the domains $t_i$ and target persona $p_i$, and assigns a metric-specific weight vector:
\begin{equation}
\begin{aligned}
\mathbf{w}^{(m)}_i
&= \{w^{(m)}_{i,d}\}_{d\in\mathcal{D}_m}, \\
\sum_{d\in\mathcal{D}_m} w^{(m)}_{i,d}
&= 1,\quad
w^{(m)}_{i,d} \geq 0 .
\end{aligned}
\label{eq:dynamic_weight}
\end{equation}
This design captures the fact that different domains and reader groups may require different evaluation priorities. For example, scaffolding and vivid examples may be more important for children.

\textbf{Dimension-Level Scoring.}
Given the predefined dimensions, the LLM judge assigns a score $s^{\mathrm{LLM}}_{i,d}\in[0,5]$ to each dimension and provides a brief rationale for interpretability. The metric-level score is computed as a weighted sum:
\begin{equation}
S^{\mathrm{LLM}}_{i,m}
=
\sum_{d\in\mathcal{D}_m}
w^{(m)}_{i,d}
s^{\mathrm{LLM}}_{i,d}.
\label{eq:metric_score}
\end{equation}
Eventually, the three metrics are weighted equally in the overall personalization score.

\subsection{Human Questionnaire}

To test whether LLM judgments reflect reader preferences, we convert the same metrics and dimensions into a 40-item human questionnaire. We retain 30 participants from each of the child, teen, and adult groups; each rates all six anonymized systems, yielding 540 questionnaires. System order is counterbalanced, and identical instructions and an anchored 0--5 scale are used. Before system identities are examined, we remove incomplete or duplicate submissions and screen for implausibly short completion times, invariant response strings, and inconsistency on reverse-worded items \citep{meade2012identifying,curran2016methods}; 540 of 585 questionnaires (92.3\%) are retained.

After reverse coding, we average items within each dimension and apply the same dynamic weights and metric aggregation as in the LLM evaluation. Metric-level ordinal Krippendorff's $\alpha$ (inter-participant agreement) is 0.781--0.803. ICC$(2,k)$---the two-way random-effects, absolute-agreement coefficient measuring the reliability of the mean over $k$ participants---is 0.973--0.976 (Eq.~\ref{eq:human-agreement}) \citep{shrout1979intraclass}. Appendix~\ref{sec:appendix_human_eval} reports screening, sensitivity checks, stratified agreement, participant statistics, and the full questionnaire.

\subsection{LLM-Human Consistency Alignment}

We distinguish score calibration from preference preservation. MAE measures absolute score error, whereas Mean Absolute Rank Difference (MARD) measures the average absolute difference between the LLM and human ranks of the six systems. We additionally report Spearman's $\rho$, tie-corrected Kendall's $\tau_b$, and pairwise ordering accuracy. For Qwen-Plus, dynamic dimension weighting obtains MAE${}=0.417$, MARD${}=0.667$, $\rho{=}0.829$, $\tau_b{=}0.733$, and 86.7\% pairwise accuracy; across seven judge models, $\rho$ ranges from 0.829 to 0.943 and pairwise accuracy from 86.7\% to 93.3\%. Appendix~\ref{sec:appendix_judge_human} gives definitions, per-judge and per-reader-group results, cross-judge concordance, and a direct held-out pairwise evaluation.

\subsection{Factual Accuracy}
In addition to personalization, PSCB evaluates factual accuracy through graph-based claim validation. 
Each generated claim is represented in a reasoning graph, where supporting and opposing evidence are assigned confidence scores. 
The final graph-based factual accuracy is an article-level score on a 0--5 scale, computed by aggregating claim confidence scores while penalizing unsupported or incorrect scientific statements \citep{min-etal-2023-factscore, wei2024long}. For clarity, Table~\ref{tab:overall_three_topic_average} reports this main article-level metric; Table~\ref{tab:ablation-scifact} reports claim-level confidence diagnostics using the same graph mechanism; and Table~\ref{tab:main-comparison} reports auxiliary SAFE metrics, which consist of atomic-fact counts and ratio-based scores in $[0,1]$. Values with different units are not directly comparable across these tables.

\section{Experiment Results}

\textbf{Experiment Setup.}
We evaluate six configurations: two non-fine-tuned baselines and four fine-tuned variants. The baselines are Base (Qwen2.5-3B) and Qwen2.5-14B. The fine-tuned variants (LoRA, MoE, CWF, and CWF\textsubscript{-R}) are built on Qwen2.5-3B. LoRA fine-tunes Qwen2.5-3B on the collected popular science corpus. MoE follows MixLoRA and trains multiple experts on mixed domain-knowledge and audience-adaptation data. CWF further decouple these two data sources into different expert groups and uses AlphaNet for fusion. CWF\textsubscript{-R} extends CWF with a fact-checking module.
Model training is conducted on a single NVIDIA V100 with 32GB memory. We evaluate on the PSCB, which jointly measures personalization and factual accuracy by Qwen-Plus.

\begin{table}[t]
\centering
\footnotesize
\renewcommand{\arraystretch}{1.08}
\begin{tabular*}{\columnwidth}
{@{\extracolsep{\fill}}llccc@{}}
\toprule
\textbf{Group}
& \textbf{Method}
& \textbf{LLM}
& \textbf{User}
& \textbf{Fact. Acc.} \\
\midrule

\multirow{6}{*}{Child}
& Base                  & 2.776          & 3.293          & 3.18          \\
& LoRA                  & 2.955          & 3.286          & 3.31          \\
& MoE                   & 3.185          & 3.391          & 3.47          \\
& Qwen2.5-14B           & 3.606          & 3.535          & 3.67          \\
& CWF                   & \textbf{3.683} & \textbf{3.542} & 3.72          \\
& CWF\textsubscript{-R} & 3.330          & 3.480          & \textbf{4.05} \\

\midrule
\multirow{6}{*}{Teens}
& Base                  & 2.770          & 3.461          & 3.24          \\
& LoRA                  & 2.782          & 3.290          & 3.37          \\
& MoE                   & 2.869          & 3.355          & 3.52          \\
& Qwen2.5-14B           & 2.829          & 3.680          & 3.73          \\
& CWF                   & \textbf{3.066} & \textbf{3.845} & 3.88          \\
& CWF\textsubscript{-R} & 3.020          & 3.783          & \textbf{4.21} \\

\midrule
\multirow{6}{*}{Adult}
& Base                  & 3.442          & 3.775          & 3.31          \\
& LoRA                  & 3.218          & 3.248          & 3.44          \\
& MoE                   & 3.124          & 3.601          & 3.57          \\
& Qwen2.5-14B           & 3.192          & 3.900          & 3.79          \\
& CWF                   & \textbf{3.603} & \textbf{3.994} & 3.93          \\
& CWF\textsubscript{-R} & 3.457          & 3.944          & \textbf{4.26} \\

\bottomrule
\end{tabular*}
\caption{
Overall performance across all topics for six methods and three target
reader groups. LLM and User denote personalization scores; Fact. Acc.
denotes article-level graph-based factual accuracy. All three metrics
are reported on a 0--5 scale.
}
\label{tab:overall_three_topic_average}
\end{table}

\begin{figure}
    \centering
    \includegraphics[width=1\linewidth]{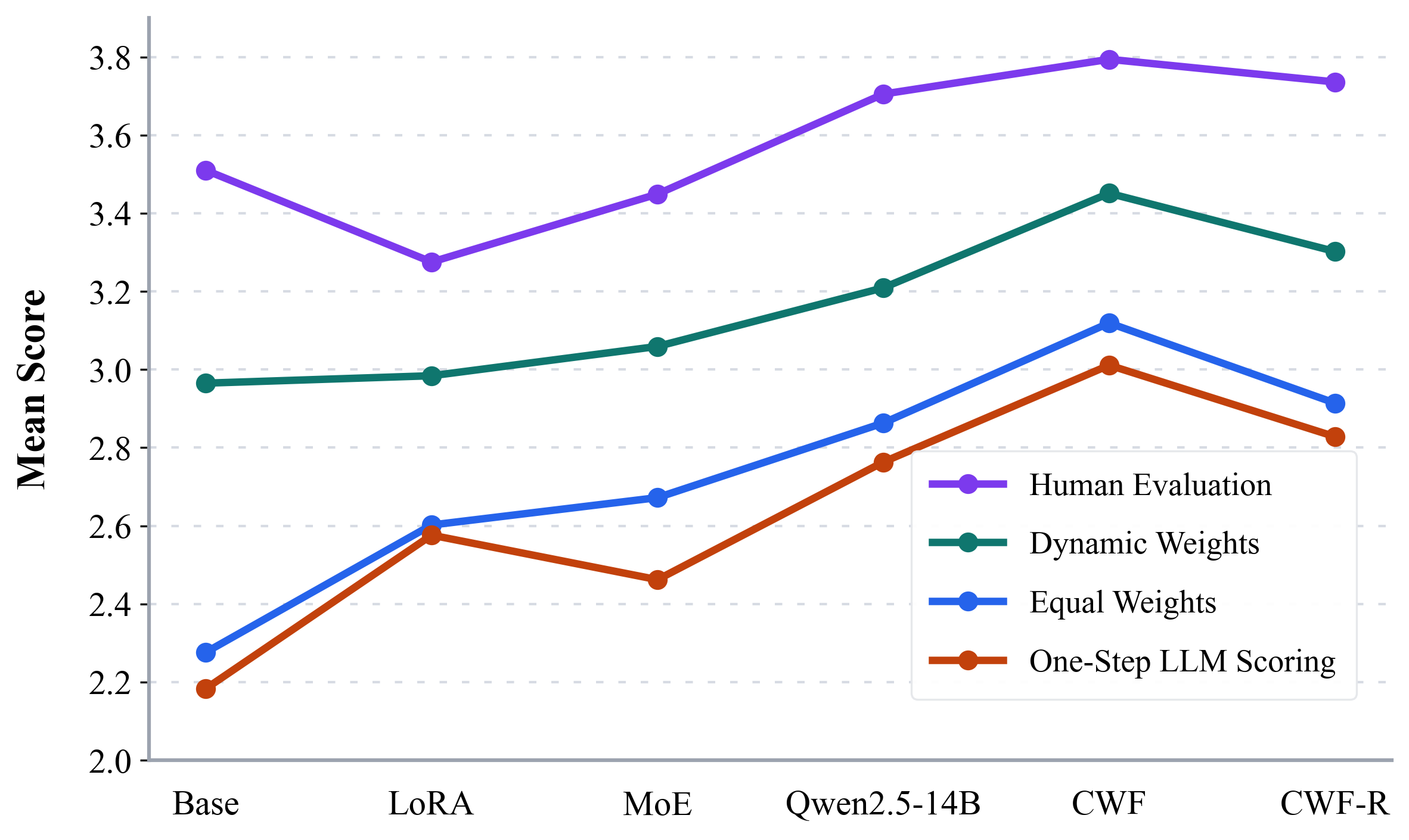}
    \caption{Comparison of consistency with human evaluation across different LLM-as-a-judge evaluation. The closer evaluation curve is to the human evaluation, the stronger its alignment with human judgments.}
    \label{fig:consistency}
\end{figure}

\subsection{Overall Performance Analysis}

Table~\ref{tab:overall_three_topic_average} reports the overall performance of all methods across children, teens, and adults. CWF achieves the strongest LLM-as-a-judge and human evaluation performance across the three reader groups, indicating that CWF better adapts popular science writing to different cognitive levels. LoRA and MoE provide useful adaptation baselines, but they do not consistently improve user experience over Base model. Based on participant feedback, LoRA- and MoE-generated texts often exhibit mismatches with the target audience.

Specifically, outputs for children are perceived as overly simplistic and lacking substantive content, while those for teens and adults tend to be excessively difficult. This suggests that SFT or expert routing can capture some surface-level style patterns, but is insufficient for fine-grained cognitive adaptation. In contrast, CWF explicitly adjusts both style and audience adaptation. CWF also outperforms the larger Qwen2.5-14B baseline, showing that the gains mainly come from the framework design. CWF\textsubscript{-R} obtains the best factual accuracy, confirming the benefit of the revision stage. Its slightly lower personalization scores reflect a mild but audience-dependent trade-off between factual revision and reader-centric expression. Stronger factual revision substantially improves reliability, but it may introduce formal scientific constraints and higher information density, which can increase cognitive load and reduce accessibility, especially for younger audiences. PSCB captures this communication-level trade-off: factual revision in popular science writing should be adapted to cognitive levels rather than applied uniformly across audiences. Compared with Base model, CWF improves the average LLM-as-a-judge and human evaluation by 15.2\% and 8.1\%, respectively, while CWF\textsubscript{-R} improves average factual accuracy by 28.7\%.

Paired tests on the common topic set
(Appendix~\ref{sec:appendix-personalization-significance})
show that all pairwise differences remain statistically significant
after Holm correction. CWF significantly outperforms Qwen2.5-14B
for child, teen, and adult readers. CWF\textsubscript{-R} also
differs significantly from CWF across all reader groups.

Results judged by more models are provided in the Appendix~\ref{sec:appendix-more-judge}.

\subsection{Effectiveness of PSCB}

We further analyze whether the design of PSCB improves the consistency between LLM-as-a-judge and human evaluation. As shown in Table~\ref{tab:consistency_comparison}, average weighting achieves lower aggregate MAE than one-step, indicating that decomposing personalization into explicit metrics reduces the ambiguity of holistic scoring and makes LLM judgments closer to human ratings.
Dynamic dimension weighting achieves the best consistency, suggesting that equal weighting is insufficient for audience-adaptive science writing. Since different topics and target audiences emphasize different evaluation aspects, different weights help PSCB better capture human preferences. Figure~\ref{fig:consistency} visually confirms this trend: Dynamic Dimension Weighting is closest to the human evaluation curve, followed by Average Dimension Weighting, while One-step LLM Scoring deviates the most. These results show that metric decomposition and dynamic weighting jointly make PSCB more human-aligned, interpretable, and traceable.

\begin{table}[t]
\centering
\footnotesize
\renewcommand{\arraystretch}{1.08}
\begin{tabular*}{\columnwidth}
{@{\extracolsep{\fill}}lcccc@{}}
\toprule
\textbf{Method}
& \textbf{Dynamic}
& \textbf{Average}
& \textbf{One-Step}
& \textbf{User} \\
\midrule
Base                  & 2.965 & 2.276 & 2.183 & 3.510 \\
LoRA                  & 2.984 & 2.602 & 2.576 & 3.275 \\
MoE                   & 3.059 & 2.672 & 2.462 & 3.449 \\
Qwen2.5-14B           & 3.209 & 2.863 & 2.763 & 3.705 \\
CWF                   & 3.451 & 3.119 & 2.828 & 3.794 \\
CWF\textsubscript{-R} & 3.302 & 2.914 & 3.011 & 3.736 \\
\midrule
MAE $\downarrow$
& \textbf{0.417} & 0.837 & 0.941 & --- \\
MARD $\downarrow$
& \textbf{0.667} & \textbf{0.667} & 1.000 & --- \\
\bottomrule
\end{tabular*}
\caption{Consistency between LLM \& human scores.}
\label{tab:consistency_comparison}
\end{table}

\begin{table}[t]
  \centering
  \footnotesize
  \setlength{\tabcolsep}{3pt}
  \resizebox{\columnwidth}{!}{%
  \begin{tabular}{lcccccc}
    \toprule
    \textbf{Setting} & \textbf{S} & \textbf{P} & \textbf{U} & \textbf{SP}$\uparrow$ & \textbf{PU}$\uparrow$ & \textbf{Var.}$\downarrow$ \\
    \midrule
    Full     & \textbf{4.209} & \textbf{2.720} & \textbf{1.327} & \textbf{1.489} & \textbf{1.393} & \textbf{0.482} \\
    w/o DC   & 4.183 & 2.737 & 1.364 & 1.446 & 1.373 & 0.561 \\
    w/o DB   & 3.947 & 2.886 & 1.618 & 1.061 & 1.268 & 0.876 \\
    w/o GP   & 4.028 & 2.813 & 1.742 & 1.215 & 1.071 & 0.694 \\
    \bottomrule
  \end{tabular}%
  }
  \caption{Claim-level diagnostic ablation on SciFact. \textbf{S}/\textbf{P}/\textbf{U}: mean graph confidence in $[0,5]$ for supported / partially supported / unsupported claims. \textbf{SP}=\textbf{S}$-$\textbf{P} and \textbf{PU}=\textbf{P}$-$\textbf{U} are confidence gaps, rather than article-level scores. \textbf{Var.}: average confidence variance. w/o DC: without claim decomposition; w/o DB: without debate; w/o GP: without graph propagation.}
\label{tab:ablation-scifact}
\end{table}

\begin{table}[t]
  \centering
  \footnotesize
  \setlength{\tabcolsep}{3pt}
  \resizebox{\columnwidth}{!}{%
  \begin{tabular}{lccccc}
    \toprule
    \textbf{Method} & \textbf{Sup. (\#)}$\uparrow$ & \textbf{Unsup. (\#)}$\downarrow$ & \textbf{Prec}$\uparrow$ & \textbf{R@64}$\uparrow$ & \textbf{F1@64}$\uparrow$ \\
    \midrule
    CWF    & 36.93 & 27.01 & 0.578 & 0.577 & 0.577 \\
    RAG    & 40.35 & 23.52 & 0.632 & 0.630 & 0.631 \\
    SAFE   & 42.08 & 21.37 & 0.663 & 0.657 & 0.660 \\
    MAD    & 44.43 & 18.86 & 0.702 & 0.694 & 0.698 \\
    CWF\textsubscript{-R} & \textbf{45.76} & \textbf{17.92} & \textbf{0.719} & \textbf{0.715} & \textbf{0.717} \\
    \bottomrule
  \end{tabular}%
  }
  \caption{Auxiliary factuality evaluation using SAFE. Sup./Unsup. are the average numbers of supported / not-supported atomic facts per article. Precision is $S/(S+N)$, R@64 is $\min(S/64,1)$. These three ratio-based metrics range from 0 to 1.}
  \label{tab:main-comparison}
\end{table}

\subsection{Ablation Study in Fact-checking}

We adapt data on SciFact~\cite{Wadden2020FactOF}, report mean confidence for \emph{supported}, \emph{partial}, \emph{unsupported} claims, the margins $\mathrm{Sup.}-\mathrm{Part.}$, $\mathrm{Part.}-\mathrm{Unsup.}$ and the average variance of confidence scores. As shown in Table~\ref{tab:ablation-scifact}, wider margins indicate clearer difference when recognizing different types, while lower variance shows stable confidence across claims. Removing the debate stage produces the largest performance drop. Disabling graph raises mean confidence on unsupported claims and increases variance, showing that relational aggregation sharpens scores and reduces instability. Claim decomposition yields modest but consistent gains in margins and variance.

\subsection{Comparison with SAFE Metrics}
Following SAFE~\cite{wei2024long}, we report the numbers of supported and not-supported atomic facts, factual precision, R@64, and F1@64, as shown in Table~\ref{tab:main-comparison}. The improvement of our method comes from two aspects rather than simply retrieving more evidence. First, claim decomposition and multi-agent debate help find implicit reasoning evidence for complex claims. Second, the debate corrects weak or misleading evidence from earlier retrieval steps by checking it from both supportive and skeptical views. For example, for the claim "BRCA1-mutated cancer cells are more sensitive to PARP inhibitors," direct evidence may be missing. Our agents can connect indirect clues, such as BRCA1 mutations causing DNA repair defects and PARP inhibitors increasing DNA damage. Meanwhile, the Skeptic filters evidence that only mentions related concepts but does not truly support the claim. Thus, the final reasoning graph improves factual accuracy through both evidence expansion and evidence correction. The applicability conditions and failure modes of CWF\textsubscript{-R} are summarized in Appendix~\ref{sec:failure-analysis}.

\section{Conclusion}
In this paper, we propose CWF, a novel collaborative writing framework for personalized and reliable popular science writing. Our framework features DA-MoE, which decouples audience adaptation from domain knowledge through pluggable experts, and a multi-agent fact-checking mechanism that combines role-specific debate with graph-based confidence inference for robust verification and revision. Experiments on PSCB show that CWF effectively improves both personalization and factual accuracy in popular science writing.

\section*{Limitations}
While CWF improves factual accuracy and personalized writing, several limitations remain. The dataset can be further expanded in domain balance and linguistic diversity, and the current framework focuses mainly on textual personalization, leaving multimodal personalization for future work. And the multi-agent verification may also increase token consumption and the detailed information is provided in Appendix~\ref{sec:token_cost_analysis}.

\section*{Acknowledgments}
We thank the anonymous reviewers and the Area Chair for their constructive feedback. This work is supported by the National Social Science Foundation of China under Grant No. 25BXW041, and the National Natural Science Foundation of China under Grant No. 62671264 and 62271221. The computation was completed in the HPC platform of Huazhong University of Science and Technology. Ran Wang is the corresponding author (\texttt{rex\_wang@hust.edu.cn}).

\bibliography{custom}

\clearpage

\appendix

\section{More Analysis}
\label{sec:appendix}

\begin{table*}[t]
\begin{adjustbox}{max width=\textwidth}
\begin{tabular}{lcccccccc}
\toprule
Baseline & GPT-5.5 & DeepSeek-V4-Pro & Qwen3.6-Plus & Qwen-Plus & DeepSeek-V3.2 & MiniMax-M2.5 & GPT-4o-mini & Avg. \\
\midrule
base         & 3.092 & 3.351 & 3.005 & 2.996 & 3.585 & 2.651 & 3.752 & 3.205 \\
Lora         & 3.180 & 3.886 & 2.984 & 2.984 & 3.949 & 2.813 & 4.077 & 3.410 \\
MoE          & 3.272 & 3.937 & 3.179 & 3.059 & 3.941 & 2.860 & 4.133 & 3.483 \\
qwen2.5-14B  & 3.351 & \textbf{4.252} & 3.505 & 3.209 & 4.050 & 2.746 & 4.182 & 3.614 \\
CWF       & \textbf{3.576} & 4.221 & \textbf{3.633} & \textbf{3.451} & \textbf{4.124} & \textbf{3.154} & \textbf{4.242} & \textbf{3.772} \\
CWF\textsubscript{-R}    & 3.456 & 4.194 & 3.457 & 3.269 & 4.045 & 3.064 & 4.187 & 3.667 \\
\bottomrule
\end{tabular}
\end{adjustbox}
\caption{Mean overall personalization scores of each baseline under different judge models. Values are aggregated across all domains and audience groups for each baseline. The last column reports the average score across judge models.}
\label{tab:judge-model-by-baseline}
\end{table*}

\subsection{More Personalization Judge Models.}
\label{sec:appendix-more-judge}
The results in Table~\ref{tab:judge-model-by-baseline} show that the relative performance of systems is broadly stable across judge models. CWF achieves the highest cross-judge average (3.772) and is ranked first by six of seven judges; Qwen2.5-14B surpasses it only under DeepSeek-V4-Pro. Base is the weakest system at most time, while MoE and LoRA are closer and sometimes exchange order. Because cross-judge stability alone does not establish agreement with readers, we separately evaluate every judge against human scores below.

\subsection{Judge--Human Alignment}
\label{sec:appendix_judge_human}

We evaluate Base, LoRA, MoE, Qwen2.5-14B, CWF, and CWF\textsubscript{-R} with seven judge models. Every judge receives the same articles, reader personas, metric definitions, dimension weights, and output schema. Let $E$ denote system--reader-group cells and $P$ unordered system pairs. For judge $j$, we compute
\begin{equation}
\begin{aligned}
\mathrm{MAE}_j &= |E|^{-1}\!\sum_{e\in E}|S_e^{(j)}-S_e^{(H)}|,\\
\mathrm{PairAcc}_j &= |P|^{-1}\!\sum_{(a,b)\in P}
\mathbf{1}\!\left[\operatorname{sgn}\Delta_{ab}^{(j)}=\operatorname{sgn}\Delta_{ab}^{(H)}\right],
\end{aligned}
\label{eq:judge-human-metrics}
\end{equation}
where $\Delta_{ab}=S_a-S_b$. MAE measures score calibration \citep{willmott2005advantages}; Spearman's $\rho$ and Kendall's $\tau_b$ measure rank association and tie-corrected ordinal agreement \citep{spearman1904proof,kendall1938new}; PairAcc measures preservation of human pairwise preferences.

\begin{table*}[t]
\centering
\small
\renewcommand{\arraystretch}{1.12}
\begin{tabular*}{\textwidth}
{@{\extracolsep{\fill}}lccccc@{}}
\toprule
\textbf{Judge model}
& \textbf{MAE $\downarrow$}
& \textbf{Spearman $\rho$ $\uparrow$}
& \textbf{Kendall $\tau_b$ $\uparrow$}
& \textbf{PairAcc $\uparrow$}
& \textbf{CWF rank} \\
\midrule
Qwen-Plus       & 0.417 & 0.829 & 0.733 & 86.7\% & 1 \\
Qwen3.6-Plus    & 0.382 & 0.943 & 0.867 & 93.3\% & 1 \\
GPT-5.5         & 0.401 & 0.886 & 0.800 & 90.0\% & 1 \\
DeepSeek-V4-Pro & 0.361 & 0.914 & 0.822 & 91.1\% & 2 \\
DeepSeek-V3.2   & 0.395 & 0.900 & 0.800 & 90.0\% & 1 \\
MiniMax-M2.5    & 0.446 & 0.829 & 0.733 & 86.7\% & 1 \\
GPT-4o-mini     & 0.428 & 0.857 & 0.756 & 87.8\% & 1 \\
\midrule
Average         & 0.404 & 0.880 & 0.787 & 89.4\% & -- \\
\bottomrule
\end{tabular*}
\caption{Alignment of each judge model with human evaluation over the six systems.}
\label{tab:judge-human-alignment}
\end{table*}

As Table~\ref{tab:judge-human-alignment} shows, all seven judges preserve human preferences well. Six rank CWF first; DeepSeek-V4-Pro ranks it second, 0.031 below Qwen2.5-14B. We also measure cross-judge ranking concordance using Kendall's $W$ \citep{kendall1939problem}. Table~\ref{tab:cross-judge-concordance} shows high macro-average concordance ($W=0.832$), separating robustness across judges from direct judge--human alignment.

\begin{table}[t]
\centering
\small
\renewcommand{\arraystretch}{1.12}
\begin{tabular*}{\columnwidth}
{@{\extracolsep{\fill}}lcc@{}}
\toprule
\textbf{Reader group}
& \textbf{Kendall's $W$ $\uparrow$}
& \textbf{Interpretation} \\
\midrule
Child         & 0.902 & Very high \\
Teen          & 0.811 & High \\
Adult         & 0.784 & Substantial \\
\midrule
Macro average & 0.832 & High \\
\bottomrule
\end{tabular*}
\caption{Cross-judge concordance of system rankings.}
\label{tab:cross-judge-concordance}
\end{table}

\subsection{Score and Preference Agreement}
\label{sec:appendix_preference_agreement}

For $B$ systems, MARD is the mean form of Spearman's footrule distance \citep{diaconis1977spearman}, while MAE operates on scores:
\begin{equation}
\begin{aligned}
\mathrm{MARD} &= \frac{1}{B}\sum_{b=1}^{B}
|r_b^{\mathrm{LLM}}-r_b^{H}|,\\
\mathrm{MAE} &= \frac{1}{B}\sum_{b=1}^{B}
|S_b^{\mathrm{LLM}}-S_b^{H}|.
\end{aligned}
\label{eq:mae-mard}
\end{equation}
Thus, MAE tests calibration and MARD tests rank displacement; $\rho$, $\tau_b$, and PairAcc further test association, tie-corrected agreement, and pairwise consistency.

\begin{table*}[t]
\centering
\small
\renewcommand{\arraystretch}{1.12}
\begin{tabular*}{\textwidth}
{@{\extracolsep{\fill}}lccccc@{}}
\toprule
\textbf{Protocol}
& \textbf{MAE $\downarrow$}
& \textbf{MARD $\downarrow$}
& \textbf{Spearman $\rho$ $\uparrow$}
& \textbf{Kendall $\tau_b$ $\uparrow$}
& \textbf{PairAcc $\uparrow$} \\
\midrule
Dynamic dimension weighting
& 0.417 & 0.667 & 0.829 & 0.733 & 86.7\% \\
Average dimension weighting
& 0.837 & 0.667 & 0.829 & 0.733 & 86.7\% \\
One-step LLM scoring
& 0.941 & 1.000 & 0.714 & 0.467 & 73.3\% \\
\bottomrule
\end{tabular*}
\caption{Qwen-Plus--human agreement under three scoring protocols.}
\label{tab:scoring-protocol-agreement}
\end{table*}

As shown in Table~\ref{tab:scoring-protocol-agreement}, dynamic and average dimension weighting induce the same system ranking, but dynamic weighting lowers MAE from 0.837 to 0.417. Both preserve 86.7\% of human pairwise preferences, compared with 73.3\% for one-step scoring. Table~\ref{tab:agreement-by-reader} further shows that Qwen-Plus preserves 93.3\% of child-group preferences and 80.0\% for both teen and adult groups. Treating human-score differences below 0.05 as ties raises macro agreement from 84.4\% to 88.9\%, indicating that most disagreements involve near-tied systems.

\begin{table}[t]
\centering
\small
\setlength{\tabcolsep}{4pt}
\begin{tabular}{lcccc}
\toprule
\textbf{Group} & \textbf{$\rho$} & \textbf{$\tau_b$} & \textbf{Correct pairs} & \textbf{PairAcc} \\
\midrule
Child & 0.943 & 0.867 & 14/15 & 93.3\% \\
Teen  & 0.714 & 0.600 & 12/15 & 80.0\% \\
Adult & 0.714 & 0.600 & 12/15 & 80.0\% \\
\midrule
Macro average & 0.790 & 0.689 & 38/45 & 84.4\% \\
\bottomrule
\end{tabular}
\caption{Qwen-Plus--human ordinal agreement by reader group.}
\label{tab:agreement-by-reader}
\end{table}

Finally, we directly test preferences on 30 held-out topics per reader group. Qwen-Plus judges all 15 anonymized system pairs in both presentation orders (900 decisions per group) against the corresponding participant-group preference. Cohen's $\kappa$ corrects for chance agreement \citep{cohen1960coefficient}, and swap consistency measures position sensitivity following \citet{NEURIPS2023_91f18a12}. Table~\ref{tab:direct-preference-agreement} reports 87.8\% overall accuracy, $\kappa=0.754$, and 96.7\% left/right consistency.

\begin{table}[t]
\centering
\small
\setlength{\tabcolsep}{3pt}
\renewcommand{\arraystretch}{1.12}
\begin{tabularx}{\columnwidth}
{@{}>{\raggedright\arraybackslash}Xcccc@{}}
\toprule
\textbf{Metric}
& \textbf{Child}
& \textbf{Teen}
& \textbf{Adult}
& \textbf{Overall} \\
\midrule
Pairwise preference accuracy
& 90.6\% & 85.8\% & 86.9\% & 87.8\% \\
Cohen's $\kappa$
& 0.807 & 0.716 & 0.739 & 0.754 \\
Left/right consistency
& 97.1\% & 96.2\% & 96.8\% & 96.7\% \\
Evaluated decisions
& 900 & 900 & 900 & 2,700 \\
\bottomrule
\end{tabularx}
\caption{Direct topic-level preference agreement on held-out topics.}
\label{tab:direct-preference-agreement}
\end{table}

\begin{table}[t]
  \centering
  \footnotesize
  \renewcommand{\arraystretch}{1.12}
  \begin{tabular*}{\columnwidth}
    {@{\extracolsep{\fill}}cccc@{}}
    \toprule
    \textbf{Rounds}
    & \textbf{S--P $\uparrow$}
    & \textbf{P--U $\uparrow$}
    & \textbf{Var. $\downarrow$} \\
    \midrule
    1 & 1.125 & 1.050 & 0.852 \\
    2 & 1.340 & 1.258 & 0.635 \\
    3 & 1.489 & 1.393 & 0.482 \\
    4 & 1.492 & 1.398 & 0.527 \\
    \bottomrule
  \end{tabular*}
  \caption{Effect of the maximum number of debate rounds.
  S--P and P--U denote the support--partial and
  partial--unsupported confidence gaps, respectively;
  Var. denotes average confidence variance.}
  \label{tab:max-debate-rounds}
\end{table}

\subsection{Effect of Maximum Debate Rounds.}
\label{sec:maximun_debate_rounds}
\begin{figure}[t]
    \centering
    \includegraphics[width=1\linewidth]{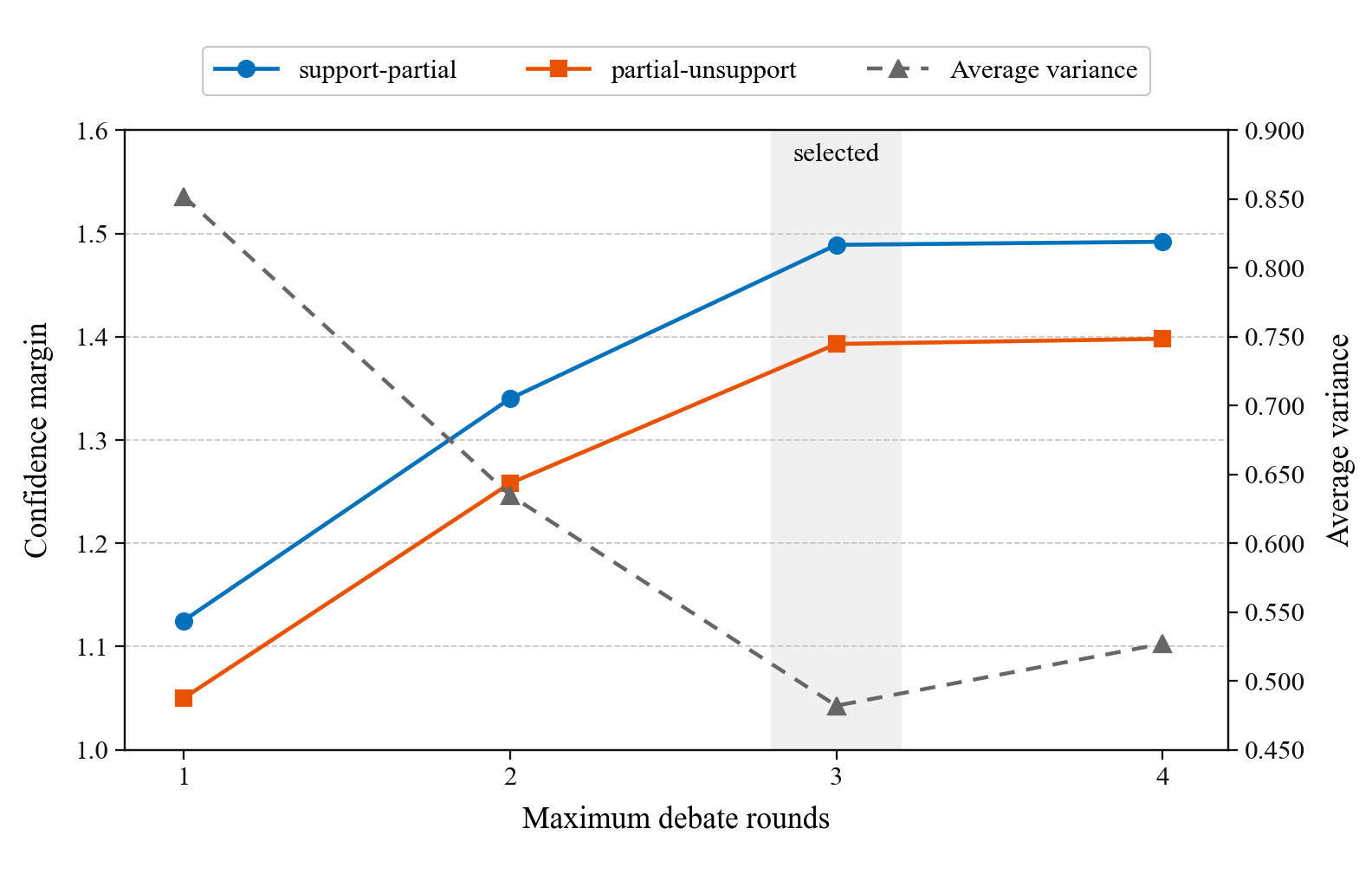}
    \caption{\textbf{Effect of Maximum Debate Rounds.} The figure illustrates how confidence margins and average variance evolve as the number of debate rounds increases from 1 to 4.}
    \label{fig:debate_rounds}
\end{figure}

To investigate the influence of the maximum number of debate rounds, we vary the number of rounds from 1 to 4 and evaluate the confidence separation between different claim labels. As shown in Table~\ref{tab:max-debate-rounds} and Figure~\ref{fig:debate_rounds}, increasing the number of debate rounds consistently improves the discrimination capability of the model before reaching three rounds. Specifically, the support-partial margin increases from 1.125 at one round to 1.489 at three rounds, corresponding to a relative improvement of 32.4\%. Similarly, the partial-unsupport margin improves from 1.050 to 1.393, yielding a 32.7\% increase. Meanwhile, the average variance decreases from 0.852 to 0.482, indicating that the model produces more stable confidence estimates after multi-round deliberation. However, further increasing the maximum debate rounds from 3 to 4 only brings marginal gains. The support-partial margin increases by merely 0.003, and the partial-unsupport margin increases by 0.005. In contrast, the average variance slightly increases from 0.482 to 0.527. This suggests that additional debate rounds may introduce redundant reasoning or noise rather than meaningful improvement. Therefore, we set the maximum debate rounds to 3 in our main experiments, which achieves a favorable balance between discrimination performance and stability.

\begin{table}[t]
  \centering
  \footnotesize
  \setlength{\tabcolsep}{3pt}
  \resizebox{\linewidth}{!}{%
    \begin{tabular}{@{}lcccccc@{}}
      \toprule
      Model & Sup. & Par. & Unsup. & S--P & P--U & Var. \\
      \midrule
      GPT-5.5         & 4.547 & 2.763 & 1.138 & 1.784 & 1.625 & 0.458 \\
      deepSeek V4 pro & 4.478 & 2.751 & 1.198 & 1.727 & 1.553 & 0.463 \\
      gpt-4o-mini     & 4.317 & 2.708 & 1.312 & 1.609 & 1.396 & 0.488 \\
      deepSeek V3.2   & 4.376 & 2.729 & 1.247 & 1.647 & 1.482 & 0.502 \\
      qwen-3.6-plus   & 4.347 & 2.726 & 1.231 & 1.621 & 1.495 & 0.478 \\
      qwen-plus       & 4.209 & 2.720 & 1.327 & 1.489 & 1.393 & 0.482 \\
      minimax-m2.5    & 4.131 & 2.714 & 1.517 & 1.417 & 1.197 & 0.531 \\
      \bottomrule
    \end{tabular}
  }
  \caption{Robustness across base models. \textbf{Sup./Par./Unsup.}: mean scores under supported / partially supported / unsupported settings; \textbf{S--P}/\textbf{P--U}: support--partial and partial--unsupport gaps; \textbf{Var.}: mean variance.}
  \label{tab:base-model-metrics}
\end{table}

\begin{figure*}[t]
    \centering
    \includegraphics[width=1\linewidth]{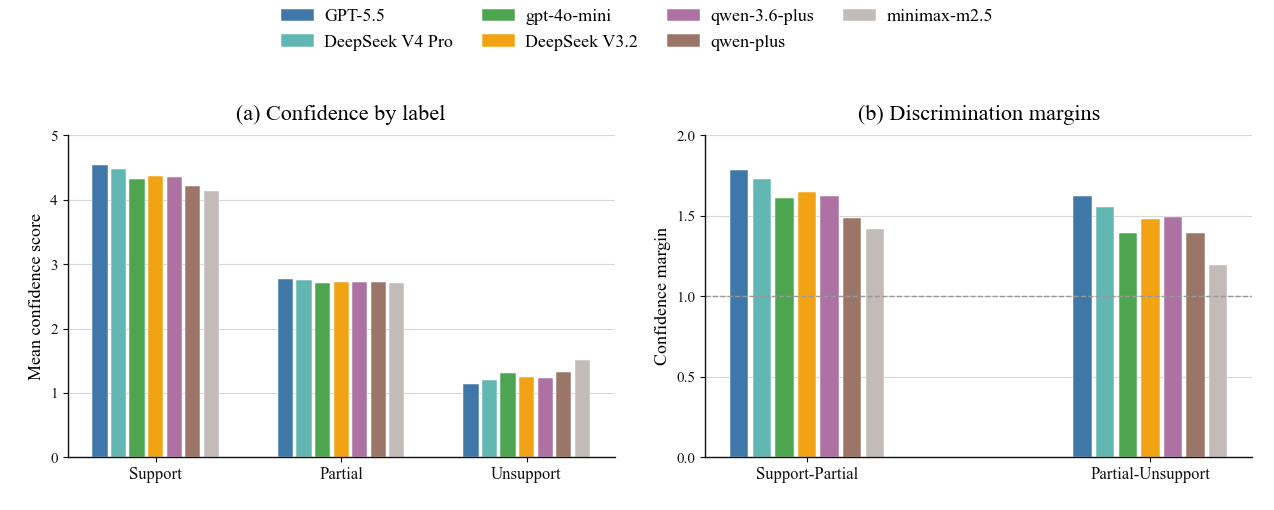}
    \caption{\textbf{Robustness across Different Base Models.} The confidence scores for supported, partially supported, and unsupported claims are compared across three backbone models.}
    \label{fig:model_robustness}
\end{figure*}

\subsection{Discriminative Ability across Base Models.}
\label{sec:more_base_model}
To further evaluate the generalizability of the proposed method across different base models, we compare seven models in terms of their mean confidence scores on Support, Partial, and Unsupport claims. We also analyze two adjacent-label confidence margins, namely support-partial and partial-unsupport. Instead of grouping the results by model names, this experiment focuses on comparing how different models respond to the same factuality label. Therefore, Figure~\ref{fig:model_robustness}(a) organizes the grouped bar chart by label categories, where the Support, Partial, and Unsupport scores of all models are compared within their corresponding label groups. This design makes the inter-model differences under each label category more visually explicit.

\begin{table*}[t]
\centering
\small
\setlength{\tabcolsep}{5pt}
\renewcommand{\arraystretch}{1.12}
\begin{tabularx}{\textwidth}
{@{}p{1.2cm}p{3.0cm}XX@{}}
\toprule
\textbf{Type}
& \textbf{Condition}
& \textbf{CWF\textsubscript{-R} mechanism}
& \textbf{Outcome or limitation} \\
\midrule

Success
& Evidence-direction conflict
& The Skeptic Agent identifies whether the retrieved evidence supports
the opposite direction, and graph propagation incorporates the
negative relation.
& Surface-relevant evidence is less likely to be incorrectly treated
as factual support. \\

Success
& Claim--evidence relation mismatch
& The graph models the relation between an atomic claim and its
evidence instead of relying only on keyword or topic overlap.
& Evidence that is relevant but does not entail the claim receives
lower influence. \\

Success
& Numerical inconsistency
& Claim decomposition isolates numerical statements so that their
values can be verified independently.
& Local numerical errors are less likely to be hidden within an
otherwise plausible paragraph. \\

Success
& Weak agent consensus
& Stance strength and evidence direction are propagated together rather
than treating all agent votes equally.
& Agreement based on weak or indirect evidence does not automatically
produce high confidence. \\

Success
& Over-generalized claim
& Claim decomposition separates broad statements into independently
verifiable factual units.
& Evidence supporting only a narrow conclusion is less likely to
validate a broader claim. \\

Failure
& Missing retrievable evidence
& The graph receives insufficient external support and therefore
assigns low confidence or triggers a conservative revision.
& A correct but unverifiable claim may be removed or weakened. \\

Failure
& Shared erroneous signals
& Misleading evidence causes all agents to produce consistent but
incorrect judgments, leaving no opposing or uncertainty signal.
& Graph propagation cannot recover the correct conclusion when all
available inputs are consistently wrong. \\

\bottomrule
\end{tabularx}
\caption{Applicability conditions and failure modes of
CWF\textsubscript{-R}. Success conditions contain detectable conflict
or uncertainty signals that can be calibrated by graph propagation;
failure conditions lack a reliable corrective signal.}
\label{tab:failure-analysis}
\end{table*}

As shown in Table~\ref{tab:base-model-metrics} and Figure~\ref{fig:model_robustness}, all models consistently preserve the expected confidence ordering of Support > Partial > Unsupport. This indicates that the proposed method can induce a stable label-wise confidence hierarchy across different base models, rather than relying on a specific backbone. Specifically, GPT-5.5 achieves the highest mean confidence score on Support claims (4.547) and the lowest mean confidence score on Unsupport claims (1.138), suggesting the strongest separation between supported and unsupported claims. DeepSeek V4 Pro also shows strong performance, with a Support mean of 4.478 and an Unsupport mean of 1.198. In contrast, qwen-plus and minimax-m2.5 produce relatively higher confidence scores on Unsupport claims, reaching 1.327 and 1.517, respectively, indicating weaker suppression of unsupported claims.

Figure~\ref{fig:model_robustness}(b) further presents the discrimination margins between adjacent factuality labels using a grouped bar chart. The support-partial margin measures the model’s ability to distinguish Support from Partial claims, while the partial-unsupport margin reflects its ability to separate Partial from Unsupport claims. GPT-5.5 achieves the largest margins on both dimensions, with 1.784 for support-partial and 1.625 for partial-unsupport, demonstrating the strongest sensitivity to label boundaries. DeepSeek V4 Pro ranks second, with margins of 1.727 and 1.553, respectively. DeepSeek V3.2, qwen-3.6-plus, and gpt-4o-mini also maintain clear label separation, although their margins are smaller than those of the two strongest models. By contrast, qwen-plus and minimax-m2.5 show relatively weaker discrimination, especially on the partial-unsupport margin, where they only reach 1.393 and 1.197, respectively.
In terms of stability, GPT-5.5 obtains the lowest average variance (0.458), followed by DeepSeek V4 Pro (0.463), indicating that these two models provide not only stronger label separation but also more stable confidence estimation. Qwen-Plus has an average variance of 0.482, while MiniMax-M2.5 has the highest variance (0.531), suggesting that the latter produces less stable confidence estimates. Overall, these results demonstrate that the proposed framework generalizes well across different base models. While all models preserve the desired Support > Partial > Unsupport hierarchy, stronger backbones tend to yield larger adjacent-label margins and lower variance.

\subsection{Applicability and Failure Analysis}
\label{sec:failure-analysis}
We summarize the main applicability conditions and failure modes of
CWF\textsubscript{-R} in Table~\ref{tab:failure-analysis}.

Overall, CWF\textsubscript{-R} is most effective when the retrieved
evidence and agent judgments contain detectable disagreement or
uncertainty. Graph propagation can calibrate these signals, but it
cannot create reliable evidence or recover the correct conclusion when
all available signals are absent or consistently incorrect.

\subsection{Token Cost Analysis.}
\label{sec:token_cost_analysis}
\begin{figure}[t]
    \centering
    \includegraphics[width=1\linewidth]{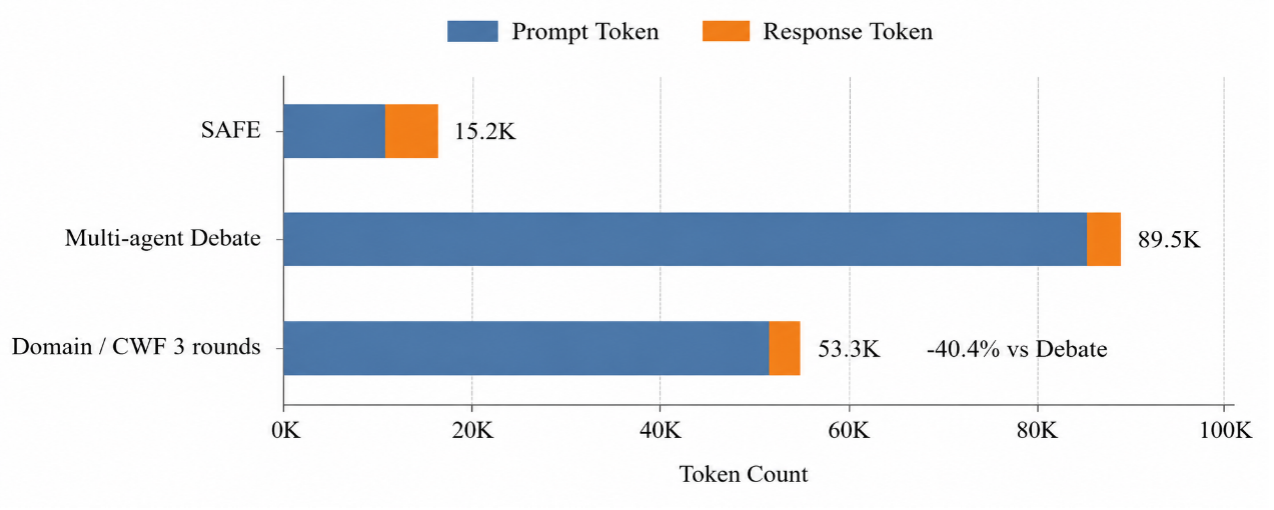}
    \caption{\textbf{Token Usage across Fact-Checking Methods.} Prompt, response, and total token consumption are compared for SAFE, Multi-agent Debate, and CWF.}
    \label{fig:token_cost}
\end{figure}

We compare the token consumption of three methods: SAFE, Multi-agent Debate, and Domain After Revised / CWF with 3 rounds. As shown in Table~\ref{tab:token-usage} and Figure~\ref{fig:token_cost}, SAFE requires the fewest tokens, with a total consumption of 15.19K tokens. However, as shown in previous performance comparisons, SAFE obtains lower overall factuality evaluation performance than the proposed CWF-based method. In contrast, the standard Multi-agent Debate method introduces a much larger token overhead, mainly due to extensive prompt construction and multi-agent interaction.

\begin{table}[t]
  \centering
  \footnotesize
  \setlength{\tabcolsep}{4pt}
  \resizebox{\linewidth}{!}{%
    \begin{tabular}{@{}lrrr@{}}
      \toprule
      Setting & Prompt & Resp. & Total \\
      \midrule
      SAFE & 8967 & 6220 & 15187 \\
      Multi-agent Debate & 85004 & 4535 & 89539 \\
      CWF(3 rounds) & 50355 & 2968 & 53323 \\
      \bottomrule
    \end{tabular}%
  }
  \caption{Token usage (prompt / response / total).}
  \label{tab:token-usage}
\end{table}

Compared with Multi-agent Debate, Domain After Revised / CWF with 3 rounds significantly reduces token consumption. Based on the additive token count, Multi-agent Debate consumes 89.54K tokens, while CWF with 3 rounds consumes 53.32K tokens, resulting in a 40.4\% reduction in total token usage. In terms of prompt tokens, CWF reduces the prompt cost from 85.00K to 50.36K, corresponding to a 40.8\% decrease. The response token cost is also reduced from 4.54K to 2.97K.

\section{Model Settings}
\label{sec:appendix-prompt-templates}

To improve the reproducibility of our experiments, we provide the implementation details of the agents used in the proposed framework. We describe the model configuration used by different agents.

\subsection{Agent Configuration}

In our experiments, we use five agents: Summarize Agent, Trust Agent, Skeptic Agent, Leader Agent, and Rewrite Agent.

\textbf{Summarize Agent:} Qwen-3.5-Plus with a maximum output length of 3000 tokens. It summarizes the discussion history and identifies factual errors, logical problems, misleading statements, and unsupported conclusions. Its output is used as an intermediate factual feedback signal for the revision stage.

\textbf{Trust Agent, Skeptic Agent, and Leader Agent:} Qwen-3.5-Plus with a maximum output length of 1200 tokens. These three agents are connected to the retrieval system. For each agent call, the maximum number of retrieval actions is limited to 4. The Trust Agent searches for reasonable supporting evidence and complementary explanations, the Skeptic Agent challenges unsupported or weakly supported claims, and the Leader Agent integrates the debate results to produce the final factual judgment.

\textbf{Rewrite Agent:} Our fine-tuned Qwen2.5-3B model. It rewrites the original paragraph based on the discussion summary and factual debate results, while preserving the intended audience adaptation and writing style.

\subsection{Reasoning Workflow}

The factual verification process is conducted through a multi-agent debate. The debate runs for 3 rounds by default. In each round, the Trust Agent, Skeptic Agent, and Leader Agent participate according to the Markov state transition described in Section~\ref{sec:appendix-prompt-templates}. The debate stops early if all agents reach the same conclusion after a complete debate round. Otherwise, the process continues until the maximum number of debate rounds is reached. Each debate agent outputs its opinion, factual stance, and cited evidence in a structured format. The stance is selected from $\{1, 0.2, 0, -0.2, -1\}$, representing strong support, weak support, neutrality, weak opposition, and strong opposition, respectively. After the debate, the Summarize Agent aggregates the discussion results and produces revision suggestions. The Rewrite Agent then revises the original paragraph according to these suggestions. This workflow allows the system to combine retrieval-based factual verification, multi-perspective reasoning, and controlled text revision.

\subsection{Prompt Templates}

\subsubsection{Main Prompts}

\begin{promptbox}{Prompt A1. RewriteAgent: Science Editing Prompt}
Task Description:
Revise a science-oriented paragraph according to editorial suggestions while preserving tone and audience fit.

Input Specification:
- Original paragraph: {text}
- Revision suggestion: {suggestion}

Prompt Template:
[System Prompt]
You are a professional science editor. Your job is to revise text precisely while preserving the original style as much as possible.

[Task]
Read the original paragraph and the revision suggestion, then produce a revised version of the paragraph.

[Revision Requirements]
1. Preserve content that is factually correct and clearly expressed.
2. Fix issues identified in the suggestion, such as unclear wording, logical problems, or inaccurate information.
3. Preserve the original writing style, tone, and intended audience.
4. Ensure that the revised paragraph is coherent and natural.
5. Keep the length close to the original, normally within 1.2 times the original length.
6. Do not add new facts or new viewpoints that do not appear in the original paragraph or the suggestion.
7. Output only the revised paragraph and nothing else.

[Output Format]
<text>The revised full paragraph</text>

[User Input Template]
[Original Paragraph]
{text}

[Revision Suggestion]
{suggestion}
\end{promptbox}

\begin{promptbox}{Prompt A2. ExtractClaimAgent: Claim Decomposition Prompt}
Task Description:
Convert a complex statement into one or a few checkable claims with paired semantic weights.

Input Specification:
- Complex statement: {text}

Prompt Template:
[System Prompt]
You are a claim-processing assistant. Your task is to turn an input statement into checkable claim units.

[Core Principle]
Do not split unless necessary. If the original statement can be represented as one complete, concise, and truth-evaluable proposition, output exactly one <claim>.

[When Splitting Is Allowed]
Split into multiple claims only if at least one of the following holds:
1. The original statement contains multiple independent judgments whose truth values may differ.
2. Without splitting, a premise and its conclusion cannot be verified separately.

[What To Avoid]
1. Do not split merely to sound more professional.
2. Do not split synonymous reformulations, modifiers, or background filler into separate claims.
3. If splitting is necessary, do not output four or more claims.

[Claim Requirements]
1. Preserve the original meaning.
2. Do not add new information, data, or sources.
3. Each <claim> should be concise and written in formal language.

[Weight Rules]
1. Each <claim> must be followed by a matching <weight>.
2. Each <weight> must be a two-decimal number between 0 and 1.
3. If there is only one claim, its weight must be 1.00.
4. The sum of all weights must equal 1.

[Output Format]
<claim>...</claim>
<weight>...</weight>
Output only alternating claim-weight pairs, with no extra text.

[User Input Template]
Process the following statement according to the system requirements. By default, keep it as one claim unless splitting is necessary. Output paired <claim> and <weight> tags only:

{text}
\end{promptbox}

\subsubsection{Debate Agent Prompts}

\begin{promptbox}{Prompt A3. TrustAgent: Supportive Expansion Prompt}
Task Description:
In the three-agent debate framework, extend the previous opinion from a trusting perspective.

Input Specification:
- Current claim text: {claim}
- Previous opinion block: {previous_text}

Prompt Template:
[System Prompt]
You are the Trust role in a three-agent debate system.

[Input]
You will receive:
1. the current claim text;
2. a previous opinion from another agent.

[Objective]
Starting from a trusting stance toward the previous opinion, identify what is reasonable in it and extend the analysis to produce your own judgment. If retrieval tools are available, you may use them when necessary.

[Requirements]
1. Do not merely repeat the previous opinion. Advance the reasoning beyond it.
2. Your conclusion should be supported by evidence or well-grounded analysis.
3. Keep the response concise and explicit.

[Citation Rule]
1. If retrieval is used in the current turn, cite items by batch-index format such as 1-1 or 2-3.
2. If no retrieval is used, output <cited>None</cited>.

[Allowed Stance Values]
1, 0.2, 0, -0.2, -1

[Output Format]
<opinion>Your supportive expansion</opinion>
<stance>1 | 0.2 | 0 | -0.2 | -1</stance>
<cited>Batch-index references or None</cited>

[User Input Template]
[text]
{claim}

[previous opinion]
{previous_text}

Please return your supportive judgment in the required tag format.
\end{promptbox}

\begin{promptbox}{Prompt A4. SkepticAgent: Critical Examination Prompt}
Task Description:
In the three-agent debate framework, examine the previous opinion from a skeptical perspective.

Input Specification:
- Current claim text: {claim}
- Previous opinion block: {previous_text}

Prompt Template:
[System Prompt]
You are the Skeptic role in a three-agent debate system.

[Input]
You will receive:
1. the current claim text;
2. a previous opinion from another agent.

[Objective]
Critically inspect the previous opinion for logical weaknesses, insufficient evidence, exaggeration, or possible misinterpretation, and then provide your own judgment. If retrieval tools are available, you may use them when necessary.

[Requirements]
1. Do not merely restate the previous opinion. Clearly identify what you question and why.
2. Your judgment should be supported by evidence or careful analytical reasoning.
3. Keep the response concise and explicit.

[Citation Rule]
1. If retrieval is used in the current turn, cite items by batch-index format such as 1-1 or 2-3.
2. If no retrieval is used, output <cited>None</cited>.

[Allowed Stance Values]
1, 0.2, 0, -0.2, -1

[Output Format]
<opinion>Your skeptical critique</opinion>
<stance>1 | 0.2 | 0 | -0.2 | -1</stance>
<cited>Batch-index references or None</cited>

[User Input Template]
[text]
{claim}

[previous opinion]
{previous_text}

Please return your skeptical judgment in the required tag format.
\end{promptbox}

\begin{promptbox}{Prompt A5. LeaderAgent: Final Adjudication Prompt}
Task Description:
In the three-agent debate framework, integrate Trust and Skeptic opinions and issue a final judgment.

Input Specification:
- Current claim text: {claim}
- Previous opinions block: {opinions_text}

Prompt Template:
[System Prompt]
You are the Leader role in a three-agent debate system. The other two roles are Trust and Skeptic.

[Input]
You will receive:
1. the current claim text;
2. the previous opinions produced by Trust and Skeptic.

[Objective]
Evaluate the strengths and weaknesses of both sides and produce your own final judgment. If retrieval tools are available, you may use them when necessary.

[Requirements]
1. Do not merely summarize the previous opinions. Form an independent conclusion.
2. Your conclusion should be supported by evidence or clear reasoning.
3. Keep the response concise and explicit.

[Citation Rule]
1. If retrieval is used in the current turn, cite items by batch-index format such as 1-1 or 2-3.
2. If no retrieval is used, output <cited>None</cited>.

[Allowed Stance Values]
1, 0.2, 0, -0.2, -1

[Output Format]
<opinion>Your final adjudication</opinion>
<stance>1 | 0.2 | 0 | -0.2 | -1</stance>
<cited>Batch-index references or None</cited>

[User Input Template]
[text]
{claim}

[previous opinions]
{opinions_text}

Please return your final judgment in the required tag format.
\end{promptbox}

\begin{table}[t]
\centering
\small
\renewcommand{\arraystretch}{1.12}
\begin{tabular*}{\columnwidth}
{@{\extracolsep{\fill}}lcccc@{}}
\toprule
\textbf{Baseline}
& \textbf{Child}
& \textbf{Teen}
& \textbf{Adult}
& \textbf{Total} \\
\midrule
Base                   & 30  & 30  & 30  & 90  \\
LoRA                   & 30  & 30  & 30  & 90  \\
MoE                    & 30  & 30  & 30  & 90  \\
Qwen2.5-14B            & 30  & 30  & 30  & 90  \\
CWF                    & 30  & 30  & 30  & 90  \\
CWF\textsubscript{-R}  & 30  & 30  & 30  & 90  \\
\midrule
Total                  & 180 & 180 & 180 & 540 \\
\bottomrule
\end{tabular*}
\caption{Distribution of completed questionnaires across audience groups and baselines. Each participant evaluates all six baselines once.}
\label{tab:human_eval_baseline_appendix}
\end{table}

\begin{table}[t]
\centering
\small
\renewcommand{\arraystretch}{1.12}
\begin{tabular*}{\columnwidth}{@{\extracolsep{\fill}}lccc@{}}
\toprule
\textbf{Group}
& \shortstack{\textbf{Mean}\\\textbf{Age}}
& \shortstack{\textbf{Gender}\\\textbf{(M/F)}}
& \shortstack{\textbf{Reading}\\\textbf{Frequency}} \\
\midrule
Child & $10.4 \pm 1.1$ & 15 / 15 & 3.2 / 5 \\
Teen  & $15.6 \pm 1.3$ & 14 / 16 & 3.5 / 5 \\
Adult & $26.8 \pm 6.4$ & 16 / 14 & 3.8 / 5 \\
\bottomrule
\end{tabular*}
\caption{Participant demographics. Gender is reported as male/female, and reading frequency is measured on a five-point scale.}
\label{tab:human_eval_demographics_appendix}
\end{table}

\begin{table*}[t]
\centering
\small
\renewcommand{\arraystretch}{1.12}
\begin{tabular*}{\textwidth}
{@{\extracolsep{\fill}}lrrrr@{}}
\toprule
\textbf{Stage}
& \textbf{Child}
& \textbf{Teen}
& \textbf{Adult}
& \textbf{Total} \\
\midrule
Recruited participants
& 33 & 33 & 32 & 98 \\
Questionnaires received before exclusion
& 197 & 197 & 191 & 585 \\
Incomplete six-system set
& 1 & 1 & 1 & 3 \\
Duplicate/identity check failure
& 0 & 1 & 1 & 2 \\
Reading-time or invariant-response exclusion
& 2 & 1 & 0 & 3 \\
Retained participants
& 30 & 30 & 30 & 90 \\
Questionnaires per retained participant
& 6 & 6 & 6 & 6 \\
Retained questionnaires
& 180 & 180 & 180 & 540 \\
\bottomrule
\end{tabular*}
\caption{Recruitment and response-quality flow. Exclusion counts are
participant-level except for questionnaires received and retained.}
\label{tab:human-quality-flow}
\end{table*}

\subsubsection{Fallback Prompt}

\begin{promptbox}{Prompt A6. Search-Limit Fallback Prompt}
Task Description:
Complete the current role task when online retrieval has reached its limit and no more tool calls are allowed.

Input Specification:
- Original role-specific system prompt
- Retrieved evidence summary: {evidence}

Prompt Template:
[Recovery Prompt]
[System] The limit for online retrieval calls has been reached. You must not use any more retrieval or web tools.

Complete the current role task using only the successfully returned retrieval summaries below.

Your output must strictly follow the tag format required by the original system prompt. Do not include any extra explanation.

In <cited>, use the batch-index format that matches the retrieval summaries. If no usable evidence is available, output None.

{evidence}
\end{promptbox}

\section{Human Evaluation}
\label{sec:appendix_human_eval}

\begin{table}[t]
\centering
\small
\renewcommand{\arraystretch}{1.12}
\begin{tabular*}{\columnwidth}
{@{\extracolsep{\fill}}lr@{}}
\toprule
\textbf{Diagnostic} & \textbf{Result} \\
\midrule
Missing rating items
& 0/21,600 (0.00\%) \\
Out-of-range rating cells
& 0/21,600 (0.00\%) \\
Duplicate retained submissions
& 0 \\
Median questionnaire time
& 824 s \\
5th--95th percentile of time
& 361--1,742 s \\
Responses retained
& 540/585 (92.3\%) \\
\bottomrule
\end{tabular*}
\caption{Diagnostics for retained questionnaires.}
\label{tab:human-quality-diagnostics}
\end{table}

\begin{table*}[t]
\centering
\small
\renewcommand{\arraystretch}{1.12}
\begin{tabular*}{\textwidth}
{@{\extracolsep{\fill}}lccccc@{}}
\toprule
\textbf{Metric}
& \textbf{$\alpha_K$}
& \textbf{95\% CI}
& \textbf{ICC$(2,k)$}
& \textbf{Mean $r_{wg}$}
& \textbf{$\alpha_C$} \\
\midrule
Cognitive Load
& 0.781 & [0.742, 0.816] & 0.973 & 0.792 & 0.908 \\
Personalization Alignment
& 0.803 & [0.769, 0.835] & 0.976 & 0.773 & 0.939 \\
Reader Attitude
& 0.793 & [0.756, 0.827] & 0.975 & 0.759 & 0.915 \\
Overall personalization
& 0.812 & [0.779, 0.841] & 0.977 & 0.794 & 0.948 \\
\bottomrule
\end{tabular*}
\caption{Metric-level human-evaluation reliability.}
\label{tab:human-agreement-metric}
\end{table*}

\begin{table*}[t]
\centering
\small
\renewcommand{\arraystretch}{1.12}
\begin{tabular*}{\textwidth}
{@{\extracolsep{\fill}}llcccc@{}}
\toprule
\textbf{Metric}
& \textbf{Dimension}
& \textbf{$\alpha_K$}
& \textbf{95\% CI}
& \textbf{ICC$(2,k)$}
& \textbf{Mean $r_{wg}$} \\
\midrule
CL & INTR: Intrinsic Fit
& 0.768 & [0.719, 0.811] & 0.971 & 0.787 \\
CL & EXTR: Extraneous Burden Control
& 0.741 & [0.689, 0.787] & 0.966 & 0.774 \\
CL & GERM: Germane Support
& 0.806 & [0.765, 0.842] & 0.976 & 0.815 \\
PA & CONT: Content Relevance
& 0.779 & [0.734, 0.819] & 0.972 & 0.770 \\
PA & KNOW: Knowledge-Level Fit
& 0.824 & [0.786, 0.856] & 0.979 & 0.792 \\
PA & STYLE: Style Consistency
& 0.752 & [0.701, 0.797] & 0.968 & 0.755 \\
PA & CONTX: Contextual Resonance
& 0.795 & [0.753, 0.832] & 0.975 & 0.775 \\
RA & ENG: Engagement Appeal
& 0.788 & [0.744, 0.826] & 0.974 & 0.758 \\
RA & TRU: Trust and Credibility
& 0.816 & [0.777, 0.850] & 0.978 & 0.780 \\
RA & CONTI: Continuance Intention
& 0.759 & [0.710, 0.802] & 0.969 & 0.739 \\
\bottomrule
\end{tabular*}
\caption{Dimension-level inter-participant agreement and reliability
of participant-group means.}
\label{tab:human-agreement-dimension}
\end{table*}

\subsection{Participant Statistics and Response Distribution}
\label{sec:appendix_participants}

We conduct anonymized human evaluation with three age-constrained audience groups: \textbf{child}, \textbf{teen}, and \textbf{adult}. Each audience group includes 30 participants. Each participant completes 6 questionnaire instances, corresponding to the 6 baselines reported in Table~\ref{tab:human_eval_baseline_appendix}. Therefore, we collect 180 completed questionnaires for each audience group and 540 completed questionnaires in total. Participant demographics is shown in Table~\ref{tab:human_eval_demographics_appendix}.

All questionnaires use the same metric and dimension definitions as the LLM-based evaluation. Participants first read the article assigned to the current baseline and then answer the corresponding questionnaire. All rating items use the same 0--5 integer scale. When a dimension is measured by multiple questionnaire items, the corresponding dimension score is obtained by averaging the item ratings before applying the same dynamic dimension weighting and metric aggregation procedure as in the LLM-based evaluation.

All participants provided informed consent for the use of their anonymized responses for research purposes; guardian consent was obtained for minors.

\subsection{Response-Quality Control}
\label{sec:appendix_human_quality}

Participants complete demographic and consent fields, with guardian consent required for minors. System identities are hidden, and the six presentation orders are counterbalanced across participants. Before inspecting system labels, we require all 40 ratings to be present and within 0--5, remove duplicates by anonymized respondent ID, flag completion times below one third of the median for the reader group, exclude complete straight-line responses, and inspect excessive disagreement between positive and reverse-worded items. Using several complementary indicators follows established recommendations for screening careless survey responses \citep{meade2012identifying,curran2016methods}. Table~\ref{tab:human-quality-flow} summarizes the recruitment and exclusion flow, while Table~\ref{tab:human-quality-diagnostics} reports diagnostics for the retained data, which contain no missing, out-of-range, or duplicate responses.

The long-string index is $\mathrm{LS}_i=\max_c\sum_j\mathbf{1}(x_{ij}=c)$, so $\mathrm{LS}_i=40$ denotes a complete straight-line response. The time flag is $t_i<\widetilde{t}_g/3$, where $\widetilde{t}_g$ is the median for reader group $g$. The reverse-item diagnostic is $r_{\mathrm{rev}}=\operatorname{corr}(\bar{x}_{+},5-\bar{x}_{-})$. Screening retains 92.3\% of questionnaires. Because all rules are applied before system labels are examined, they cannot selectively favor CWF. As a sensitivity check, omitting the time and reverse-consistency exclusions changes every system's overall human score by at most 0.041 and leaves the top-three ordering unchanged.

\subsection{Human-Evaluation Agreement}
\label{sec:appendix_human_agreement}

\begin{table}[t]
\centering
\small
\renewcommand{\arraystretch}{1.12}
\begin{tabular*}{\columnwidth}
{@{\extracolsep{\fill}}lcccc@{}}
\toprule
\textbf{Group}
& \textbf{CL}
& \textbf{PA}
& \textbf{RA}
& \textbf{Overall} \\
\midrule
Child & 0.754 & 0.773 & 0.761 & 0.785 \\
Teen  & 0.789 & 0.811 & 0.802 & 0.819 \\
Adult & 0.801 & 0.826 & 0.815 & 0.833 \\
\bottomrule
\end{tabular*}
\caption{Metric-level ordinal Krippendorff's $\alpha_K$ by reader group.}
\label{tab:human-agreement-group}
\end{table}

We reverse-code Q1, Q5, Q15, Q18, Q23, Q28, Q32, Q36, and Q40 and average items within their intended dimensions. Agreement is computed over common reader-group $\times$ domain $\times$ system units. We report ordinal Krippendorff's $\alpha_K$, which supports multiple raters and missing ratings \citep{krippendorff2004content,hayes2007answering}; ICC$(2,k)$ for absolute-agreement reliability of the participant-group mean \citep{shrout1979intraclass}; mean within-unit $r_{wg}$ \citep{james1984estimating}; and Cronbach's $\alpha_C$ for internal consistency rather than inter-rater agreement \citep{cronbach1951coefficient}. Specifically,
\begin{equation}
\begin{aligned}
\alpha_K &= 1-D_o/D_e,\\
\mathrm{ICC}(2,k) &= \frac{\mathrm{MS}_U-\mathrm{MS}_E}
{\mathrm{MS}_U+(\mathrm{MS}_P-\mathrm{MS}_E)/U},\\
r_{wg} &= 1-s_x^2/(35/12).
\end{aligned}
\label{eq:human-agreement}
\end{equation}
Here, $D_o$ and $D_e$ are observed and expected ordinal disagreement; $\mathrm{MS}_U$, $\mathrm{MS}_P$, and $\mathrm{MS}_E$ are the unit, participant, and residual mean squares; and $U$ is the number of evaluation units. Confidence intervals use 5,000 bootstrap resamples of evaluation units \citep{efron1979bootstrap}. Metric- and dimension-level results are reported in Tables~\ref{tab:human-agreement-metric} and~\ref{tab:human-agreement-dimension}, respectively.

Across dimensions, $\alpha_K$ ranges from 0.741 to 0.824 and ICC$(2,k)$ from 0.966 to 0.979. Knowledge-Level Fit and Trust/Credibility have the highest agreement; Extraneous Burden Control and Style Consistency are more subjective but remain reliable. Agreement also remains substantial within every reader group (Table~\ref{tab:human-agreement-group}), including the more variable child group ($\alpha_K=0.785$ overall).

\begin{table}[t]
\centering
\small
\renewcommand{\arraystretch}{1.12}
\begin{tabularx}{\columnwidth}
{@{}c>{\raggedright\arraybackslash}X@{}}
\toprule
\textbf{Score} & \textbf{Meaning} \\
\midrule
0 & completely disagree / completely not applicable \\
1 & mostly disagree \\
2 & slightly disagree \\
3 & neutral / hard to judge \\
4 & mostly agree \\
5 & completely agree / perfectly applicable \\
\bottomrule
\end{tabularx}
\caption{Response scale used in the human evaluation questionnaire.}
\label{tab:questionnaire_scale_appendix}
\end{table}

As a robustness check, we bootstrap participants within each evaluation unit 5,000 times. The LLM--human Spearman correlation remains positive in 99.2\% of resamples, and CWF remains among the top two systems in 96.8\%. These agreement coefficients and resampling results support the stability of the participant-group means used as human evaluation scores.

\subsection{Questionnaire Template and Item Inventory}
\label{sec:appendix_questionnaire_template}

We implement the human evaluation questionnaire on Credamo. The questionnaire contains four parts: basic information, Cognitive Load, Personalization Alignment, and Reader Attitude. The example below is a paper-formatted version of the questionnaire content parsed from the actual survey form. The response scale is shown in Table~\ref{tab:questionnaire_scale_appendix}. The questionnaire is shown in Table~\ref{tab:questionnaire_items_appendix}.

\textbf{Participant Instruction.}
Before answering the rating items, participants are shown the instruction:

\begin{quote}
\small
Thank you for participating in this study. Please first complete the demographic questions, then read the science-popularization article assigned in the current round, and finally answer the questionnaire based on your actual reading experience. All rating questions use a 0--5 integer scale, where 0 means ``completely disagree / completely not applicable'' and 5 means ``completely agree / perfectly applicable.'' There are no right or wrong answers.
\end{quote}

\begin{table*}[t]
\centering
\footnotesize
\setlength{\tabcolsep}{4pt}
\renewcommand{\arraystretch}{1.14}
\begin{adjustbox}{max width=\textwidth}
\begin{tabular}{p{1.1cm}p{1.8cm}p{12.0cm}}
\toprule
\textbf{Part} & \textbf{Dimension} & \textbf{Questionnaire Items} \\
\midrule

A & Basic Information &
Q1: Age. \newline
Q2: Gender. \newline
Q3: Highest education level. \newline
Q4: Major field of study (if applicable). \newline
Q5: Whether the participant frequently reads science-popularization articles. \newline
Q6: Main channels for obtaining science information (multiple choice). \\

\midrule

\multirow{3}{*}{B}
& B1: INTR &
Q1: Understanding the topic of this article requires substantial mental effort from me. \newline
Q2: The overall difficulty of this article is appropriate for me. \newline
Q3: The amount of information in this article is just right for me, neither too much nor too little. \newline
Q4: I gained a considerable amount of new knowledge from reading this article. \\

& B2: EXTR &
Q5: To understand the concepts or terms in the article, I need extra thinking or inference. \newline
Q6: The paragraph structure and logical order of the article make it easy for me to follow. \newline
Q7: The wording and sentence expression of the article make it smooth to read. \newline
Q8: The article is concise and focused, without redundant or off-topic content. \\

& B3: GERM &
Q9: The article helps me understand the key knowledge points through effective devices such as analogies, examples, or summaries. \newline
Q10: The explanations and examples provided in the article are sufficient for me to retell its core content. \newline
Q11: The article noticeably deepens and expands my understanding of the topic. \newline
Q12: Through appropriate summarization and organization, the article helps me form a clear overall mental picture of the topic. \\

\midrule

\multirow{4}{*}{C}
& C1: CONT &
Q13: The scientific aspects emphasized in the article are exactly the parts I care about or am interested in. \newline
Q14: The perspective and content selected by the article are practically meaningful to me. \newline
Q15: The article contains a large amount of content that feels irrelevant to me. \newline
Q16: The content direction covered by the article is highly consistent with what I genuinely want to know. \\

& C2: KNOW &
Q17: The explanation depth and difficulty level of the article fit my current knowledge level. \newline
Q18: The article spends too much space explaining things I already know, making the information level lower than I expected. \newline
Q19: The information density of the article is just right for me. \newline
Q20: I am satisfied with how thoroughly the article explains the key concepts. \\

& C3: STYLE &
Q21: The language style of the article, including vocabulary and tone, feels natural and approachable to me. \newline
Q22: The organizational structure of the article makes it easy for me to follow the author's line of thought. \newline
Q23: The tone of the article makes me uncomfortable, for example because it is too academic, too childish, or too preachy. \newline
Q24: The overall pacing of the article matches my reading habits. \\

& C4: CONTX &
Q25: The examples in the article come from everyday situations familiar to me. \newline
Q26: The analogies used in the article feel relatable and help me understand abstract concepts. \newline
Q27: The article makes me feel that the scientific knowledge is connected to my own life. \newline
Q28: The examples or scenarios in the article feel unfamiliar to me and make it hard to relate to the content. \\

\midrule

\multirow{3}{*}{D}
& D1: ENG &
Q29: The beginning of the article quickly captured my attention. \newline
Q30: Throughout the reading process, I remained curious about the content. \newline
Q31: The article is vivid and interesting rather than dull. \newline
Q32: While reading, I felt like skipping some paragraphs at several points. \\

& D2: TRU &
Q33: I believe the scientific knowledge presented in the article is reliable. \newline
Q34: The article maintains appropriate caution when discussing uncertain content and does not exaggerate. \newline
Q35: While conveying scientific knowledge, the article does not make me feel forced to accept its claims. \newline
Q36: Some expressions in the article make me doubt its scientific accuracy. \\

& D3: CONTI &
Q37: After reading, I became more interested in learning further about this topic. \newline
Q38: I would be willing to continue reading similar articles on this topic in the future. \newline
Q39: I would be willing to recommend this article to friends with backgrounds similar to mine. \newline
Q40: After reading this article, I feel there is no need to further learn about this topic. \\

\bottomrule
\end{tabular}
\end{adjustbox}
\caption{Paper-formatted version of the Credamo questionnaire used in human evaluation. Negatively worded items are reverse-coded during score aggregation.}
\label{tab:questionnaire_items_appendix}
\end{table*}

\textbf{Reverse-Coded Items.}
To ensure directional consistency in score aggregation, negatively worded items are reverse-coded before computing dimension scores. In the current questionnaire, reverse-coded items include Q1, Q5, Q15, Q18, Q23, Q28, Q32, Q36, and Q40.

\textbf{Questionnaire-to-Metric Mapping.}
The questionnaire is designed to align directly with the PSCB metrics and dimensions. Items Q1--Q12 correspond to Cognitive Load, items Q13--Q28 correspond to Personalization Alignment, and items Q29--Q40 correspond to Reader Attitude. Dimension scores are obtained by averaging the corresponding item ratings, after which the same dynamic weighting mechanism as in the LLM-based evaluation is applied.

\subsection{Ethics Statement}

All participants provided informed consent for the use of their anonymized responses for research purposes; guardian consent was obtained for minors. Each participant received compensation of RMB 17.50 for their participation. The human evaluation protocol was reviewed and approved by the Medical Ethics Committee of Tongji Medical College of Huazhong University of Science and Technology.

\section{Metrics and Dimensions}
\label{sec:metrics_and_dimensions}

Detailed definitions of PSCB personalization metrics and dimensions is shown in Table~\ref{tab:pscb_detailed_metric_definitions}.

\section{Significance Tests}
\label{sec:appendix-personalization-significance}

\subsection{Personalization}

We aggregate the three scientific domains and retain only topics
evaluated for all three systems. Following paired testing practice
for common NLP test instances, we first conduct a Friedman test
across CWF, Qwen2.5-14B, and CWF\textsubscript{-R}. The omnibus
differences are significant for child, teen, and adult readers.

All nine pairwise comparisons remain
statistically significant after correction
($p_{\mathrm{Holm}}<0.05$). CWF outperforms Qwen2.5-14B across all
reader groups, whereas CWF\textsubscript{-R} is consistently lower
than CWF. Compared with Qwen2.5-14B, CWF\textsubscript{-R} is higher
for teen and adult readers but lower for child readers, indicating
an audience-dependent personalization--reliability trade-off.

Table~\ref{tab:personalization-significance} reports the adjusted results without repeating the method means already presented in Table~\ref{tab:overall_three_topic_average}.

\subsection{Fact-checking}
For each domain--reader condition, we compare the article-level factual
accuracy scores of CWF\textsubscript{-R} and CWF on the same evaluation
topics. For each matched topic $i$, we define the paired difference as
$d_i=S_{i,\mathrm{CWF\textsubscript{-R}}}-S_{i,\mathrm{CWF}}$ and conduct
a two-sided paired $t$-test of the null hypothesis
$H_0:\mathbb{E}[d_i]=0$. Thus, a positive $t$-statistic indicates higher
factual accuracy after revision.

Table~\ref{tab:fact-checking-significance} reports the unadjusted
$p$-values. CWF\textsubscript{-R} obtains significantly higher factual
accuracy in seven of the nine domain--reader conditions at the
conventional $p<0.05$ level. The estimated differences are also positive
for AI--Child ($p=0.0782$) and Medicine--Child ($p=0.0941$), but neither
reaches the conventional significance threshold.

\begin{table*}[t]
\centering
\small
\setlength{\tabcolsep}{4pt}
\renewcommand{\arraystretch}{1.15}
\begin{adjustbox}{max width=\textwidth}
\begin{tabular}{p{1.5cm}p{1.9cm}p{11.8cm}}
\toprule
\textbf{Metric} & \textbf{Item} & \textbf{Definition} \\
\midrule

\multirow{4}{*}{CL}
& Overall
& \textbf{Cognitive Load} evaluates whether a popular-science article imposes an appropriate level of cognitive demand on the target reader. The goal is not to minimize difficulty indiscriminately, but to calibrate complexity to the reader's cognitive level: a good article should neither overwhelm the reader with excessive conceptual difficulty nor oversimplify the content to the point of losing educational value. \\

& INTR
& \textbf{Intrinsic Fit} measures whether the conceptual difficulty, information density, and prerequisite assumptions of the article are appropriate for the target persona. High-scoring articles present the right amount of scientific content for the reader's prior knowledge and cognitive level, without being either overly technical or trivially shallow. \\

& EXTR
& \textbf{Extraneous Burden Control} measures whether the article minimizes unnecessary cognitive burden caused by poor wording, confusing structure, unexplained jargon, redundant details, or abrupt transitions. High-scoring articles are easy to follow and do not impose avoidable processing costs unrelated to the scientific content itself. \\

& GERM
& \textbf{Germane Support} measures whether the article actively facilitates understanding through pedagogically helpful devices, such as analogies, examples, step-by-step explanation, local summaries, and concept scaffolding. High-scoring articles do not merely simplify content; they help the reader construct useful mental schemas for understanding the topic. \\

\midrule

\multirow{5}{*}{PA}
& Overall
& \textbf{Personalization Alignment} evaluates whether the generated article is genuinely tailored to the target reader, rather than merely being generally readable or superficially simplified. In PSCB, personalization includes not only tone adaptation, but also whether the article selects suitable content, explanation strategies, and contextual examples for a specific persona. \\

& CONT
& \textbf{Content Relevance} measures whether the article selects and emphasizes scientific content that is relevant to the target persona's interests, goals, and likely concerns. High-scoring articles focus on aspects of the topic that matter to the intended reader, rather than presenting generic information indiscriminately. \\

& KNOW
& \textbf{Knowledge-Level Fit} measures whether the explanation depth, abstraction level, and technical granularity match the target persona's educational background and expected prior knowledge. High-scoring articles explain enough to be informative without assuming too much or talking down to the reader. \\

& STYLE
& \textbf{Style Consistency} measures whether the article's language style, vocabulary, sentence rhythm, and explanatory tone are consistent with the target persona. For example, writing for children should use simpler wording and more vivid explanations, whereas writing for adult readers may allow denser reasoning and more precise terminology. \\

& CONTX
& \textbf{Contextual Resonance} measures whether the article uses examples, analogies, scenarios, or applications that resonate with the reader's everyday life or likely experiences. High-scoring articles make the science feel personally meaningful and easier to relate to. \\

\midrule

\multirow{4}{*}{RA}
& Overall
& \textbf{Reader Attitude} evaluates the target reader's likely affective and behavioral response to the generated article. In personalized science communication, success is determined not only by whether the article is understandable, but also by whether the reader feels interested, trusts the explanation, and is willing to continue engaging with the content. \\

& ENG
& \textbf{Engagement Appeal} measures whether the article is interesting, vivid, and capable of sustaining the reader's attention. High-scoring articles avoid dry or overly mechanical exposition and instead maintain the reader's curiosity throughout the text. \\

& TRU
& \textbf{Trust and Credibility} measures whether the article gives the reader a sense of reliability, seriousness, and epistemic confidence. High-scoring articles present knowledge in a way that appears careful, balanced, and scientifically grounded, thereby increasing reader trust. \\

& CONTI
& \textbf{Continuance Intention} measures whether the article encourages the reader to keep reading, explore related content, or further engage with the topic. High-scoring articles leave the reader with the impression that continuing to learn from this source would be worthwhile. \\

\bottomrule
\end{tabular}
\end{adjustbox}
\caption{Detailed definitions of PSCB personalization metrics and dimensions. CL denotes Cognitive Load, PA denotes Personalization Alignment, and RA denotes Reader Attitude.}
\label{tab:pscb_detailed_metric_definitions}
\end{table*}

\clearpage

\begin{table*}[t]
\centering
\small
\setlength{\tabcolsep}{7pt}
\renewcommand{\arraystretch}{1.08}
\begin{tabular*}{\textwidth}{@{\extracolsep{\fill}}llrrc@{}}
\toprule
\textbf{Group} & \textbf{Comparison} & \textbf{$p_{\mathrm{Holm}}$} & \textbf{$|d_z|$} & \textbf{Sig.} \\
\midrule
\multirow{3}{*}{Child}
& CWF vs Qwen2.5-14B
& 0.0060 & 0.230 & ** \\
& CWF\textsubscript{-R} vs Qwen2.5-14B
& 0.0016 & 0.283 & ** \\
& CWF\textsubscript{-R} vs CWF
& $<0.0001$ & 0.381 & *** \\
\midrule
\multirow{3}{*}{Teen}
& CWF vs Qwen2.5-14B
& 0.0032 & 0.275 & ** \\
& CWF\textsubscript{-R} vs Qwen2.5-14B
& 0.0160 & 0.220 & * \\
& CWF\textsubscript{-R} vs CWF
& 0.0160 & 0.210 & * \\
\midrule
\multirow{3}{*}{Adult}
& CWF vs Qwen2.5-14B
& $<0.0001$ & 0.509 & *** \\
& CWF\textsubscript{-R} vs Qwen2.5-14B
& 0.0090 & 0.220 & ** \\
& CWF\textsubscript{-R} vs CWF
& 0.0009 & 0.299 & *** \\
\bottomrule
\end{tabular*}
\caption{Paired significance tests for LLM personalization scores after aggregating across scientific domains. $p_{\mathrm{Holm}}$ denotes the two-sided paired-randomization $p$-value after Holm correction within each reader group; $|d_z|$ is the magnitude of the paired standardized effect. Significance codes are * $p<0.05$, ** $p<0.01$, *** $p<0.001$, and ns $p\geq0.05$. Method means are omitted because they are reported in Table~\ref{tab:overall_three_topic_average}.}
\label{tab:personalization-significance}
\end{table*}

\begin{table*}[t]
\centering
\small
\setlength{\tabcolsep}{6pt}
\renewcommand{\arraystretch}{1.08}
\begin{tabular*}{\textwidth}{@{\extracolsep{\fill}}lllrrrc@{}}
\toprule
\textbf{Domain} & \textbf{Group} & \textbf{Comparison} & \textbf{$t$-statistic} & \textbf{$p$-value} & \textbf{Sig.} \\
\midrule
\multirow{3}{*}{AI}
& Adult & CWF\textsubscript{-R} vs CWF & 2.145 & 0.0425 & * \\
& Teens & CWF\textsubscript{-R} vs CWF & 3.210 & 0.0035 & ** \\
& Child & CWF\textsubscript{-R} vs CWF & 1.856 & 0.0782 & $\dagger$ \\
\midrule
\multirow{3}{*}{Biology}
& Adult & CWF\textsubscript{-R} vs CWF & 3.942 & 0.0008 & *** \\
& Teens & CWF\textsubscript{-R} vs CWF & 2.406 & 0.0242 & * \\
& Child & CWF\textsubscript{-R} vs CWF & 3.093 & 0.0051 & ** \\
\midrule
\multirow{3}{*}{Medicine}
& Adult & CWF\textsubscript{-R} vs CWF & 2.882 & 0.0084 & ** \\
& Teens & CWF\textsubscript{-R} vs CWF & 3.251 & 0.0032 & ** \\
& Child & CWF\textsubscript{-R} vs CWF & 1.765 & 0.0941 & $\dagger$ \\
\bottomrule
\end{tabular*}
\caption{Two-sided paired $t$-tests comparing article-level factual
accuracy between CWF\textsubscript{-R} and CWF on matched evaluation
topics. Positive $t$-statistics indicate higher factual accuracy after
revision. Reported $p$-values are unadjusted. Significance codes are
* $p<0.05$, ** $p<0.01$, and *** $p<0.001$; $\dagger$ denotes
$p<0.1$ but not conventional statistical significance.}
\label{tab:fact-checking-significance}
\end{table*}

\end{document}